\documentclass[review, 1p, a4paper]{elsarticle}
\usepackage[namelimits]{amsmath} 
\usepackage{amssymb}
\usepackage{tabularx}
\usepackage{pifont}
\usepackage{subfigure}
\usepackage{setspace}
\usepackage{array}
\usepackage[table,xcdraw]{xcolor}
\usepackage[normalem]{ulem}
\useunder{\uline}{\ul}{}
\usepackage{float}
\usepackage[ruled,vlined]{algorithm2e}
\usepackage{enumitem}
\usepackage{hyperref}
\usepackage{caption}
\usepackage{dsfont}

\usepackage[switch]{lineno}

\journal{}
\begin{document}
\captionsetup[figure]{labelfont={bf},labelformat={default},labelsep=period,name={Fig.}}
\begin{frontmatter}
\title{MNN: Mixed Nearest-Neighbors for Self-Supervised Learning}
\author[a,b]{Xianzhong Long \corref{cor1}}
\ead{lxz@njupt.edu.cn}
\author[a,b]{Chen Peng}
\author[a,b]{Yun Li}
\address[a]{School of Computer Science, Nanjing University of Posts and Telecommunications, Nanjing, 210023, China}
\address[b]{Jiangsu Key Laboratory of Big Data Security and Intelligent Processing, Nanjing, 210023, China}
\cortext[cor1]{Corresponding author}
\renewcommand{\abstractname}{Executive Summary}
\begin{abstract}
In contrastive self-supervised learning, positive samples are typically drawn from the same image but in different augmented views, resulting in a relatively limited source of positive samples. An effective way to alleviate this problem is to incorporate the relationship between samples, which involves including the top-K nearest neighbors of positive samples. However, the problem of false neighbors (i.e., neighbors that do not belong to the same category as the positive sample) is an objective but often overlooked challenge due to the query of neighbor samples without supervision information. In this paper, we present a simple self-supervised learning framework called Mixed Nearest-Neighbors for Self-Supervised Learning (MNN). MNN optimizes the influence of neighbor samples on the semantics of positive samples through an intuitive weighting approach and image mixture operations.
The results demonstrate that MNN exhibits exceptional generalization performance and training efficiency on four benchmark datasets.
\end{abstract}


\end{frontmatter}

\section{Introduction}\label{1}
\begin{figure}[h!]
	\centering
    \includegraphics[width=0.65\linewidth]{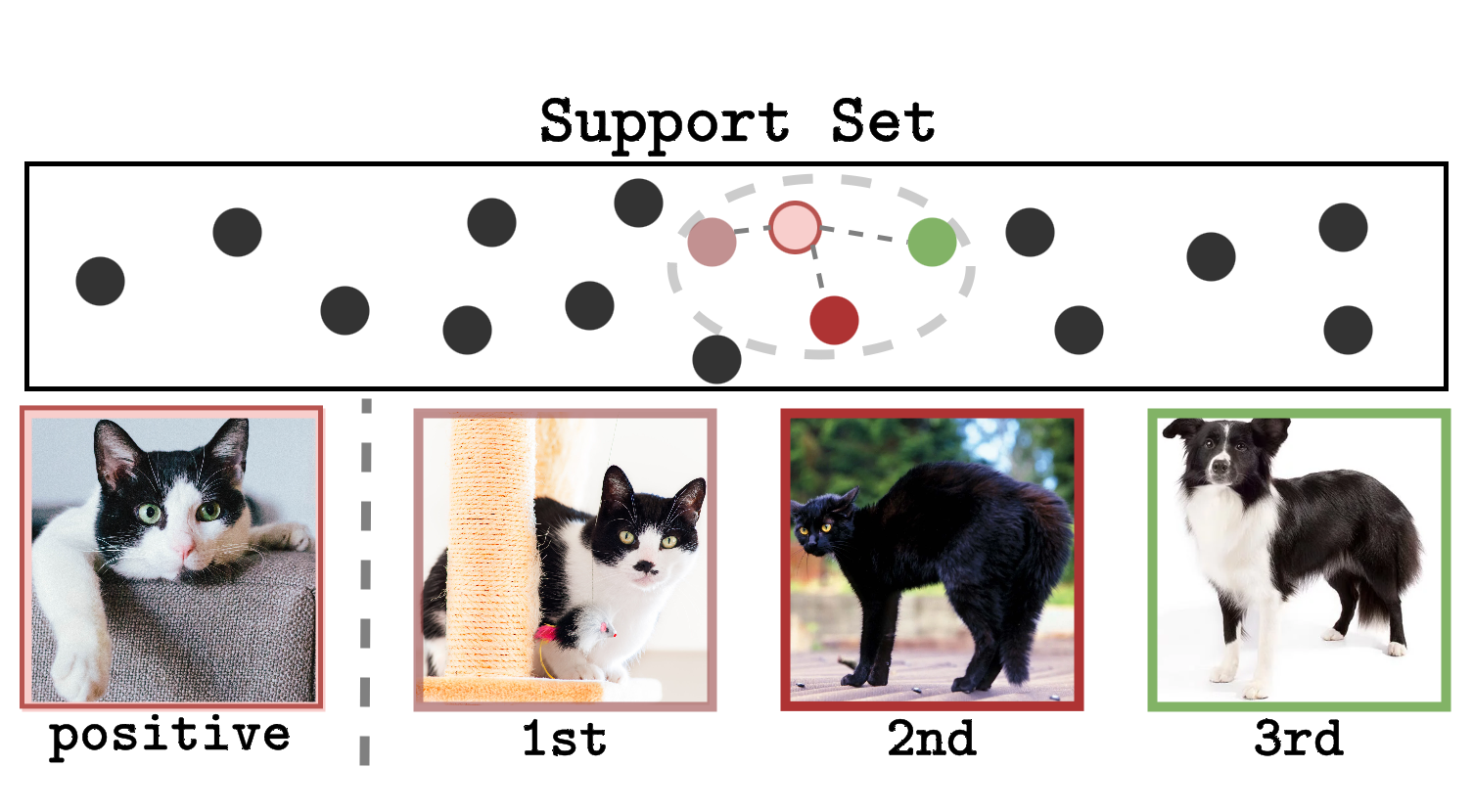}
	\centering
     \caption{The positive sample and its neighbor samples set. The set of neighbor samples is composed of the samples in the support set that have the top-K largest cosine similarity values to the positive sample (e.g., K is 3 in the figure). Samples within the neighbor set are treated as additional positive samples, thus samples that are visually similar to the positive sample but semantically different (e.g., the 3rd neighbor), which may confuse the model.}
     \label{figure: mnn_example}
     \vspace{-1.0em}
\end{figure}

Self-supervised learning (SSL) has made significant progress in the field of deep learning due to its ability to learn rich semantic features using large-scale unlabeled data \cite{simclr_2020, mocov1_2020, jigsaw_puzzles_2016, colorful_2016, context_prediction_2015}. A notable direction of this research involves instance discrimination tasks \cite{instance_discrimination_2018, contrastive_learning_2020}. Briefly, the core idea is that views of the same image with different data augmentations should be close to each other in feature space, while views of different images should be far away from each other. Some studies have incorporated nearest neighbors into contrastive self-supervised learning to address some challenges within the framework. This type of approach helps to mitigate issues like the class collision problem, which arises from neglecting intra-class relationships in the dataset. For instance, NNCLR \cite{nnclr_2021} seeks the top-K neighbors of positive samples in the support set and employs Noise Contrastive Estimation (NCE) Loss \cite{infonce_loss_2010} to align these neighbors with positive samples from another branch. MSF \cite{msf_2021} enhances the learned semantic features by identifying the top-K nearest neighbors of the corresponding positive samples in the target branch. Subsequently, MSF utilizes Mean Squared Error (MSE) Loss to connect these neighbors with the positive samples in the online branch. CMSF \cite{cmsf_2022} exploits an additional support set to enrich the semantics of neighbor samples. However, these methods ignore a significant issue: there are some samples (e.g., the 3rd neighbor in Fig.~\ref{figure: mnn_example}) in the neighbor set that are semantically inconsistent with the current instance. We term them as False Nearest-Neighbors (FNN). We alleviate this issue by mixing the positive sample and its corresponding neighbor samples in the feature space.

Inspired by SNCLR \cite{snclr_2023}, we initially considered transferring the Cross-Attention Score (CAS) from SNCLR to the MSF framework without negative samples. Although this change led to significant performance improvements in downstream tasks compared to the original MSF, further analysis (detailed in Sec. \ref{441}) revealed increased instability with CAS. As a result, we reformulated a simplified loss function, the Weighted Squared Error (WSE), to enhance method stability and interpretability while retaining the experimental performance gains.

Some previous works have also attempted to combine the mixture with SSL. For example, UnMix \cite{unmix_2022} improves the performance of SSL with additional forward propagation of mixed images. i-Mix \cite{i_mix_2021} employs mixture as a regularization strategy to increase the generalization of contrastive SSL. MoCHi \cite{mochi_2020} synthesizes more meaningful hard negative samples by mixture to facilitate model learning. Our work is different from them. Instead of mixture in pixel space, MNN first mixes the current instance and its corresponding top-K neighbors in feature space. Secondly, the mixture of MNN aims to alleviate the disturbance of false neighbors in neighbor sets, instead of treating them as hard negative samples. Finally, the additional computational overhead of the MNN can be neglected without additional forward propagation of mixed images. This enables MNN more efficient and easy to implement in applications.

Our main contributions are summarized as follows:

\begin{itemize}[itemsep=4pt,topsep=0pt,parsep=0pt]
    \item We detected instability when examining the entropy of the weight distribution associated with the Cross-Attention Score. To address this issue, we introduced a simplified loss function that effectively integrates nearest-neighbors with contrastive self-supervised learning.
    \item To mitigate the impact of false neighbors in the model, we employ a mixing strategy that combines the positive samples with their corresponding neighbor samples in the feature space. Our approach not only significantly enhances model performance with minimal additional overhead but also operates on a highly intuitive rationale.
\end{itemize}
\section{Related work}\label{2}
In this section, we will review the applications of Self-supervised learning, Nearest-neighbors exploration, and Mix-up in computer vision related to our MNN.
\subsection{Self-supervised learning }\label{21}
Self-supervised learning has made significant progress in the field of computer vision as a method for acquiring generalized semantic features without manually annotated data \cite{lgsimclr_2023, ProPos_2023, know_yourself_2023}. Early self-supervised methods were mainly based on heuristic pretext tasks (e.g., predicting image rotation \cite{rotations_2018}, recovering image color \cite{colorful_2016}). However, the performance of these methods is limited by the setting of the task.

In recent years, contrastive self-supervised learning has garnered widespread attention \cite{contrastive_learning_2020, instance_discrimination_2018, infonce_loss_2010, simsam_2021, graph_NN_2023}. Notably, methods like MoCo \cite{mocov1_2020, mocov2_2020}, SimCLR \cite{simclr_2020}, and BYOL \cite{byol_2020} have made remarkable strides in performance enhancement. They achieve this by incorporating components like momentum updates, projection heads, and asymmetric network structures to optimize the model. However, these methods still rely on data augmentation driven by prior knowledge for creating positive samples, which may limit their capacity to generalize to scenarios with insufficient prior knowledge. In this work, MNN alleviates the effect of FNN by bringing in a domain-agnostic mixture. The approach capitalizes on nearest neighbors to enhance the semantic diversity of positive samples.
\subsection{Nearest-neighbors exploration}\label{22}
The nearest-neighbor (NN) methods are widely used in computer vision tasks such as image classification \cite{knn_image_classification_2012} and domain adaptation \cite{knn_domain_adaptation_2021}. Self-supervised learning seeks to uncover nuanced relationships between samples using NN techniques. For instance, NNCLR \cite{nnclr_2021} optimizes the model's performance by identifying the top-K neighbors of positive samples and applying NCE loss. Drawing from the BYOL approach \cite{byol_2020}, MSF \cite{msf_2021} locates the nearest neighbors corresponding to positive samples and enhances their semantic features using MSE loss. However, these methods often overlook the presence of false neighbors, a potential drawback that can compromise the generalization performance of the model.

SNCLR \cite{snclr_2023} assesses the impact of distinct neighbor samples on the model using Cross-Attention Score, constructed through identity mappings and cosine similarity. While our efforts to transfer CAS to MSF \cite{msf_2021} without negative samples resulted in performance improvements, we also observed that CAS could lead to substantial model confusion. To seamlessly integrate nearest neighbors into self-supervised learning, we introduced a simplified loss function. This loss function effectively distinguishes between positive and neighbor samples by assigning distinct fixed weights.
\subsection{Mix-up}\label{23}
Mix-up \cite{mixup_2018} is a well-established regularization technique that finds applications across various learning paradigms. In essence, it trains the network by forming a convex combination of sample pairs and label pairs, thereby promoting the acquisition of semantic features with linear properties. Recent research has explored the fusion of Mix-up with self-supervised learning. For instance, MixCo \cite{mixco_2020} randomly combines two images from the same batch, enabling the model to consider the implicit relationship between positive and negative samples. i-Mix \cite{i_mix_2021} introduces image mixture as a domain-agnostic regularization approach for contrastive learning, leading to enhanced model performance in multiple modalities, such as video and speech. UnMix \cite{unmix_2022} facilitates the learning of smooth decision boundaries through self-mixtures, promoting less confident predictions. Notably, the mixing strategy employed in MNN operates within the feature space rather than the pixel space. Moreover, the mixture operation of MNN has no need for additional forward propagation, so its additional computational overhead is negligible compared to previous works.

MoCHi \cite{mochi_2020}, a related method to our work, utilizes mixture operations in the feature space to create more meaningful hard negative samples, thereby adjusting the difficulty of the pre-training task. In contrast, MNN employs mixed-sample features as additional positive samples with the specific goal of reducing noise introduced by False Nearest-Neighbors.
\begin{figure}[t]
	\centering
    \includegraphics[width=\linewidth]{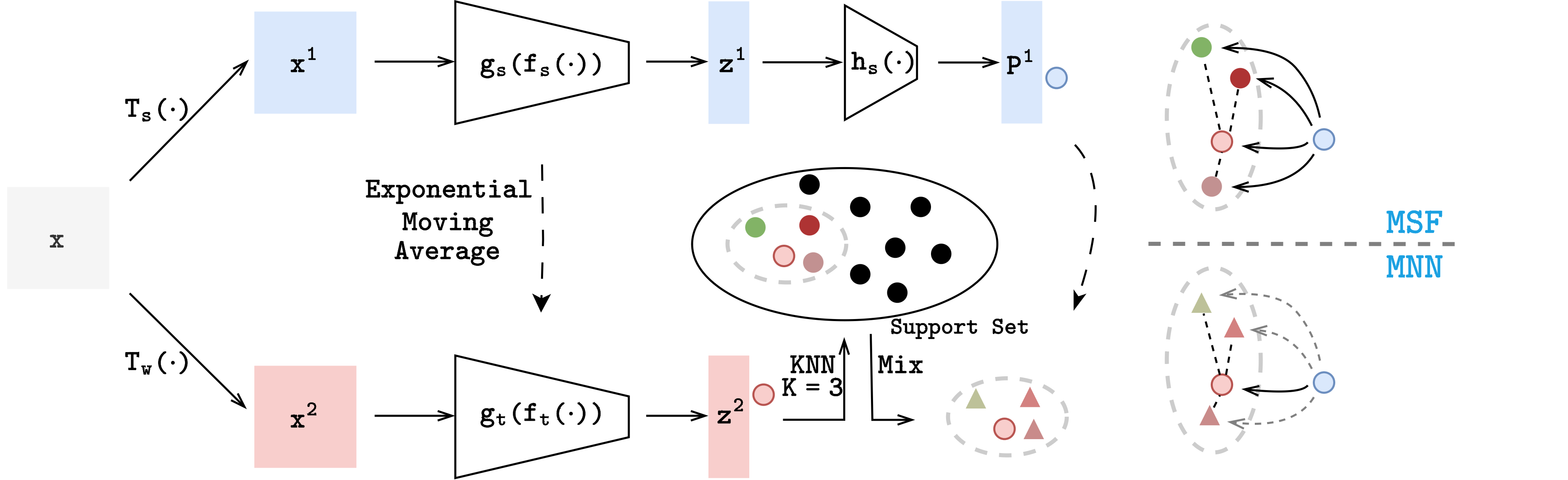}
	\centering
	
     \caption{MNN overview: We identify the top-K samples in the support set with the highest cosine similarity to $z^2$, forming the neighbor set. Each neighbor sample undergoes a mixture operation, resulting in a mixed neighbor, illustrated as a triangle. These mixed neighbors, along with $p^1$ and $z^2$ itself, contribute to the computation. The student network is updated using the loss function, while the teacher network undergoes an update through a momentum update process, which involves computing the exponential moving average of the student network. We use a straightforward loss function (WSE) and image mixture to mitigate the disturbance caused by semantically inconsistent false neighbors (e.g., the green dot in the figure) with the current instance.}
     \label{figure: diagram_mnn}
     \vspace{-1.0em}
\end{figure}
\section{Method}\label{3}
In this section, we provide a detailed explanation of our proposed MNN method, including key components such as neighbor sample querying, distinguishing between positive and neighbor samples, and image mixing.

Fig.~\ref{figure: diagram_mnn} illustrates the MNN framework. Starting with image $x$ from the current batch $X$, we apply two data augmentations to create views $x^1$ and $x^2$. These views are then processed by the student encoder and the teacher encoder, resulting in embeddings $z^1$ and $z^2$. For $z^2$, we select the top-K embeddings ${\{z^2_i\}}_{i=1}^{K}$ as its nearest neighbors based on cosine similarity with $z^2$ within the support set, which includes other sample embeddings in the dataset. Additionally, we mix these neighbor samples with $z^2$ to obtain ${\{\widetilde{z^2_i}\}}_{i=1}^{K}$. To prevent model collapse \cite{byol_2020}, we introduce a predictor $h(\cdot)$ in the student network branch, generating $p^1$. In the optimization process, we aim to minimize the following loss function:

\begin{equation}
Loss_x = {\textstyle \sum_{i=0}^{K} w_i*||p^1 - \widetilde{z_i^2}||_2^2} \ ,
\label{equation_mnn}
\end{equation}
where $\widetilde{z^2_0} = z^2$ and all embeddings are normalized before computation. The weights $w_i$ are used to determine the contribution of each ${\widetilde{z^2_i}}$ to the loss function. Finally, we update the student encoder and the teacher encoder using backpropagation and momentum mechanisms \cite{mocov1_2020}, respectively. Notably, if we exclude the mixture operation and set the weight $w_i = 1/(K+1)$, the MNN approach becomes equivalent to MSF \cite{msf_2021}. Furthermore, if we set $K=0$ and $w_0=1$, MNN degenerates to BYOL \cite{byol_2020}. Thus, MNN can be viewed as a form of generalization of the BYOL approach.
\subsection{Search for neighbor samples}\label{31}
We employ a strategy aligned with classical approaches \cite{nnclr_2021, msf_2021, snclr_2023}, initially using the support set $\mathcal{S}$ to store candidate neighbors. We rely on cosine similarity as the metric for identifying the top-K neighbor samples of $z^2$:

\begin{equation}
{\{z^2_i\}}_{i=1}^K = \underset{s\in \mathcal{S}}{argmax}(cos(z^2, s), top_n = K)\ ,
\label{equation_cosine}
\end{equation}
where ${\{z^2_i\}}_{i=1}^K$ are sorted in descending order of cosine similarity with $z^2$. The default value for K is 5, and in Sec.~\ref{432}, we conduct comparative experiments with different K values. After updating the network parameters, we employ a first-in-first-out (FIFO) strategy to refresh the batch of samples associated with the teacher network branch in $\mathcal{S}$. These strategies aim to maintain the samples in the support set, ensuring consistency and the validity of the neighbor finding and mixing processes.

In our approach, the top-K neighbor samples are treated as additional positive samples and incorporated into the Weighted Squared Error Loss along with $p^1$. Consequently, the accurate selection of these neighbor samples becomes pivotal. Since we cannot select neighbor samples based on labels in the context of SSL, the issue of False Nearest-Neighbors arises. To alleviate this, we assign different weights to distinguish the contributions of positive and neighbor samples to the model. Additionally, we mitigate the impact of FNN through image mixture.
\subsection{Distinguish between positive and neighbor samples}\label{32}
We initially tried to differentiate between positive and neighbor samples using the Cross-Attention Score \cite{snclr_2023, vit_2021}. Although this approach led to some performance improvements, our analysis revealed issues with the instability of its weight distribution (detailed in Sec.~\ref{441}). As a result, we redefined a method in which different fixed weights are assigned to distinguish the positive sample $z^2$ from the neighbor samples:

\begin{equation}
w_i^{WSE}= \begin{cases}
 1 & , \ i= 0\\
 \frac{1}{K}  & ,\text{ otherwise }.
\end{cases}
\label{equation_weight_mnn}
\end{equation}

We believe that neighbor samples can indeed augment the diversity of positive samples. However, we also acknowledge the possibility of substantial semantic distinctions between neighbor samples and true positive samples. These distinctions could potentially increase the risk of false neighbors. Consequently, it becomes essential to differentiate between neighbor samples and genuine positive samples by employing varying weights.
\subsection{Mixed images}\label{33}
Addressing the instability in weight assignment caused by CAS through the allocation of different fixed weights to positive and neighbor samples can enhance the generalization capacity of the model. However, the presence of objectively false neighbor samples, as depicted in Fig.~\ref{figure: mnn_example}, prompts us to consider methods for \textit{making these spurious neighbors more closely resemble genuine ones}. In the embedding space of the teacher network output, we utilize mix-up \cite{mixup_2018} to alleviate the disruption from FNN in the model. Specifically, for each neighbor embedding $z^2_i$, we perform a mixture operation:

\begin{equation}
\widetilde{z_i^2} = \lambda *z_i^2 + (1-\lambda)*z^2\ ,
\label{equation_mix_mnn}
\end{equation}
where the mixed coefficients $\lambda$ follow a standard uniform distribution. Our goal with this mixture approach in MNN is to divert attention away from false neighbors (for details, see \ref{A2}), thereby enhancing the focus on optimizing the relationship between positive embeddings $z^2$ and $p^1$.

\begin{table}[h]
\centering
\caption{Notations in the MNN framework.}
\resizebox{\linewidth}{!}
{
\begin{tabularx}{\textwidth}{p{0.1\textwidth}X}
\hline
Notation & \multicolumn{1}{c}{Meaning} \\ \hline 
$\mathcal{S}$ & The set consists of embeddings generated by passing samples from previous batches through the teacher encoder. The update process follows a first-in-first-out approach. \vspace*{0.4\baselineskip} \\
${\{z^2_i\}}_{i=1}^K$ & The set comprises the K neighbors of the embedding $z^2$. To identify these neighbors, we calculate the cosine similarity between $z^2$ and all embeddings in the support set $\mathcal{S}$, selecting the top-K embeddings in descending order of similarity. \vspace*{0.4\baselineskip} \\
${\{\widetilde{z^2_i}\}}_{i=1}^{K}$ & Each neighbor $\widetilde{z_i^2}$ in this set is a mixture of the embedding $z^2$ and its corresponding neighbor $z_i^2$. \vspace*{0.4\baselineskip} \\
${\{w_i\}}_{i=1}^K$ & Each element $w_i$ in the set quantifies the contribution of the neighbor embedding $\widetilde{z_i^2}$ to the loss function. \vspace*{0.4\baselineskip} \\\hline
\end{tabularx}
}
\label{table: symbols}
\end{table}

\subsection{Relation between WSE and Mixture}\label{34}
In an ideal scenario, the integration of neighbors in contrastive self-supervised learning would allow the model to consider the intra-class relationships in the dataset, thereby enhancing the semantic information of the positive sample. However, the presence of false neighbors is an inherent issue. On the one hand, WSE effectively distinguishes between the genuine positive sample and the neighbor samples by assigning them different weights. This allows the model to bring together distinct augmented views of the same image while incorporating the rich and diverse semantics offered by the neighbor samples. On the other hand, the mixture operation within MNN serves to counteract the impact of false neighbors within the neighbor set, mitigating their disruptive noise in the model. These two key components collectively enable MNN to minimize the influence of false neighbors and extract the full spectrum of rich semantic information provided by neighbor samples. We have endeavored to apply these two essential components to alternative approaches that might face issues related to false neighbors. Our findings, outlined in Sec.~\ref{443}, reveal their widespread applicability. We present the pseudo-code for MNN in \ref{A1} and provide explanations for the key symbols used in the MNN framework in Table~\ref{table: symbols}.

\begin{table}[ht]
\centering
\setlength{\belowcaptionskip}{6 pt}
\caption{Statistics of the four datasets. As the test set for Tiny ImageNet is not accessible, we conduct evaluations using the test set for all datasets except for Tiny ImageNet, for which we employ the validation set.}
\begin{tabular}{lcccc}
\hline
Dataset             & CIFAR-10  & CIFAR-100     & STL-10        & Tiny ImageNet \\ \hline
Classes             & 10        & 100           & 10            & 200           \\
Image size          & 32 x 32   & 32 x 32       & 96 x 96       & 64 x 64       \\
Training set        & 50,000    & 50,000        & 105,000       & 100,000       \\
Test set            & 10,000    & 10,000        & 8,000         & -             \\
Validation set      & -         & -             & -             & 10,000        \\ \hline
\end{tabular}
\label{table: datasets}
\end{table}

\section{Experiments}\label{4}
In this section, we evaluate the MNN framework through a series of experiments. We begin by outlining the experimental setup and then contrast MNN's performance with established self-supervised learning methods using standard evaluation protocols. Ablation studies underscore the critical components of MNN, followed by a comparison of various weight adjustment strategies within the framework. Additionally, we enrich our analysis by integrating MNN's key features into other advanced methods to evaluate the resultant performance enhancements.
\subsection{Experiment setup}\label{41}
\subsubsection{Datasets and device performance}
Although self-supervised learning methods are commonly evaluated using the ImageNet-1K dataset \cite{imagenet_2009}, conducting this type of evaluation remains challenging for many research labs due to hardware limitations. We conducted experiments on four benchmark image datasets: CIFAR-10 and CIFAR-100 \cite{cifar10_100_2009}, each containing 60,000 images, STL-10 \cite{stl10_2011} with 113,000 images, and Tiny ImageNet \cite{tiny_imagenet_2015} with 120,000 images (see Table~\ref{table: datasets}). All algorithms for this work were pre-trained and evaluated on the same hardware device (1 Nvidia GTX 3090 GPU).

\subsubsection{Data augmentation}
Data augmentation is a pivotal aspect of contrastive self-supervised learning \cite{simclr_2020}. Our approach aligns with the data augmentation strategies of MoCoV2 \cite{mocov2_2020} to generate distorted views, including random horizontal flips and color distortions. To efficiently identify the corresponding set of neighbor samples within the teacher network, preserving the semantic information of $z^2$ is of utmost importance. Consequently, we adopt weak data augmentation techniques in the teacher network, conducive to neighbor queries. Conversely, we apply strong data augmentation in the student network to ensure the learned semantic features are robust against geometric transformations in images. This strategy is denoted as '(s/w)', with additional details provided in Table~\ref{table: data_augmentation}.

\begin{table}[ht]
\centering
\setlength{\belowcaptionskip}{6 pt}
\caption{Data augmentation of the experiment. We employ 'strong' data augmentation for the student network, which aligns with typical contrastive self-supervised learning practices \cite{mocov2_2020}. For the teacher network, we use 'weak' data augmentation consisting only of Resized Crops and Horizontal Flip. This choice allows us to meet the data augmentation requirements of contrastive learning while mitigating the risk of False Nearest-Neighbors when searching for neighbors.}
\resizebox{0.9\linewidth}{!}
{
\begin{tabular}{lccccc}
\hline
       & Resized Crops & Horizontal Flip & Color Jitter & GrayScale & Gaussian Blur \\ \hline
Strong & \ding{51}     & \ding{51}       & \ding{51}    & \ding{51} & \ding{51}  \\
Weak & \ding{51}     & \ding{51}       &              &           &            \\
\hline
\end{tabular}
}
\label{table: data_augmentation}
\vspace{-1.0em}
\end{table}

\subsubsection{Network architecture}
In the encoder network, we adopt the ResNet18 \cite{resnet_2016}, which comprises a backbone $f(\cdot)$ and a projection head $g(\cdot)$. The projection head is designed with two Fully-Connected (FC) layers; the first layer has a configuration of [512, 2048], and the second [2048, 128]. Between these layers, normalization and a Rectified Linear Unit (ReLU) are implemented for non-linear processing. The prediction head $h(\cdot)$ within the student network is akin to the projection head but differs in the first layer, which has dimensions [128, 2048].

\begin{table}[ht]
\centering
\setlength{\belowcaptionskip}{6 pt}
\caption{Parameters for Pretraining and Downstream experiments. To guarantee controlled comparisons in our experiments, we applied uniform hyperparameters across all algorithms we reproduced, wherever feasible.}
\begin{tabular}{lccc}
\hline
& CIFAR-10 (CIFAR-100) & STL-10 (Tiny ImageNet) \\ \hline
\textbf{Pretraining task}   &                        &               \\ \hline
Epoch                    & 200       & 200           \\
Top-K                     & 5         & 5           \\
momentum                & 0.99       & 0.996          \\
Batch size               & 256      & 256           \\
Warm up epoch            & 5        & 5             \\
Base learning rate       & 0.06     & 0.06          \\
Support set               & 4096     & 16384         \\
Weight decay             & 5e-4     & 5e-4          \\ \hline
\textbf{Downstream task} &          &               \\ \hline
Epoch                    & 100      & 100           \\
Learning rate            & 30       & 30           \\
Weight decay             & 0        & 0             \\ \hline
\end{tabular}
\label{table: parameters}
\end{table}
\subsubsection{Pre-training}
To ensure equitable comparisons, we maintained consistent hyperparameter settings across all algorithms. Our training process involved iterative updates with an Stochastic Gradient Descent (SGD) optimizer using a momentum of 0.9 and a weight decay of 5e-4, spanning a total of 200 epochs. In the initial 5 epochs, we employed a linear warm-up strategy to gradually raise the learning rate to $\text{lr} = 0.06 \times \text{BatchSize}/256$ before transitioning to the cosine annealing schedule \cite{cosine_warm_2016}. In tandem with this, we harnessed the momentum update mechanism to optimize the teacher network parameters, denoted as $F_t = m*F_t + (1-m)*F_s$, where $m$ represents the momentum coefficient. To maintain a balance between modeling data distribution and efficiently updating sample features, the size of the support set was assigned based on the training set size in different datasets. By default, all algorithms used $K=5$ as the number of neighbors. More details are provided in Table~\ref{table: parameters}.

\subsubsection{Evaluation protocols}
We conducted linear evaluation and K-nearest neighbors (KNN) classification, which are commonly employed in SSL. During the linear evaluation phase, we initially freeze the backbone parameters of the student network. Subsequently, a linear classifier [512, \textit{cla}] is attached to the backbone of the student network, where \textit{cla} denotes the number of semantic classes relevant to the downstream task. The classifier was trained with a learning rate of 30, a weight decay of 0, and a momentum of 0.9. Learning rate adjustments were made by reducing it by a factor of 0.1 at the 60th and 80th epochs, with training lasting for a total of 100 epochs. Given the sensitivity of linear evaluation to hyperparameters, we opted for a straightforward KNN classification approach \cite{instance_discrimination_2018}. First, we preserved the parameters of the pre-trained model and transformed all the training dataset samples into 512-dimensional semantic embeddings, which were stored. During the testing phase, KNN classification was used to choose the top-K samples from the stored training samples and assign labels to the test samples based on a majority vote.

\subsection{Main results}\label{42}
\subsubsection{Linear evaluation}
\begin{table}[t]
\centering
\setlength{\belowcaptionskip}{6 pt}
\caption{Linear evaluation results. We present the top-1 accuracy of classical approaches on four benchmark image datasets. Models are categorized based on whether they require nearest neighbors for training. The best performance is highlighted in bold, while suboptimal results are underlined. Results marked with $^*$ denote those reproduced using the official code as they are not directly provided. Notably, the results for SimCLR and BYOL are taken from \cite{ressl_2021}, while the results for SCE are derived from \cite{sce_2023}. We aligned our experimental settings as closely as possible with theirs \cite{ressl_2021, sce_2023} to ensure a fair comparison.}
\begin{tabular}{lcccc}
\hline
Method                          & CIFAR-10       & CIFAR-100    & STL-10    & Tiny ImageNet  \\ \hline
Supervised \cite{ressl_2021}                     & 94.22          & 74.66        & 82.55     & 59.26          \\ \hline
SimCLR \cite{simclr_2020}       & 84.92          & 59.28        & 85.48     & 44.38          \\
BYOL \cite{byol_2020}           & 85.82          & 57.75        & 87.45     & 42.70          \\ 
SCE \cite{sce_2023}             & 90.34          & {\ul65.45}   & 89.94     & \textbf{51.90}  \\
MoCoV2$^*$ \cite{mocov2_2020}   & 89.56          & 62.47        & 88.91     & 46.38          \\
UnMix$^*$ \cite{unmix_2022}     & 90.37          & 65.30        & 90.51     & 47.29          \\ 
\hline
NNCLR$^*$ \cite{nnclr_2021}     & 87.72          & 59.62        & 87.13     & 41.52          \\
SNCLR$^*$ \cite{snclr_2023}     & 88.86          & 65.19        & {\ul90.93}     & 50.15    \\
CMSF$^*$ \cite{cmsf_2022}       & {\ul91.00}     & 62.37        & 88.21     & 44.50         \\
MSF$^*$ \cite{msf_2021}         & 89.94          & 59.94        & 88.05     & 42.68          \\ 
\rowcolor[HTML]{ECF4FF} 
MNN                  & \textbf{91.47} & \textbf{67.56}   & \textbf{91.61}     & {\ul50.70} \\ \hline
\end{tabular}
\label{table: linear_eval_main}
\end{table}
Table~\ref{table: linear_eval_main} presents the top-1 classification results obtained through linear evaluation. To ensure fair comparisons, we applied symmetric backpropagation uniformly across all algorithms. The results demonstrate that, in most cases, MNN outperforms other classical self-supervised methods. A significant distinction between MNN and MSF is the additional mixture operation in the embedding space, as discussed in Sec.~\ref{33}. Crucially, MNN significantly outperformed MSF across all four datasets, with improvements of 1.53\%, 7.62\%, 3.56\%, and 8.02\%, respectively. This underscores the effectiveness of MNN in mitigating the noise disturbance caused by False Nearest-Neighbors by working with a mixture of positive samples and neighbor samples.

\subsubsection{K-nearest neighbors}
Linear evaluations have clearly demonstrated that MNN significantly outperforms most classical methods. To validate robustness against hyperparameters during evaluation, we employed a K-nearest neighbors classifier ($K = 200$) to assess pre-trained features. As seen in Table~\ref{table: knn_main}, our method also surpasses previous classical methods in utilizing off-the-shelf features. Significantly, on this evaluation metric, MNN achieves a performance improvement over MSF by 1.57\%, 9.64\%, 2.56\%, and 6.99\% across the four datasets, respectively.

\begin{table}[ht]
\centering
\setlength{\belowcaptionskip}{6 pt}
\caption{The results of K-nearest neighbors classification ($K = 200$). Models are categorized based on whether they require nearest neighbors during training. Results labeled with $^*$ indicate reproduction using official code due to the lack of direct results. The best performance is shown in bold, while suboptimal results are underlined.}
\begin{tabular}{lcccc}
\hline
Method                          & CIFAR-10       & CIFAR-100     & STL-10       & Tiny ImageNet  \\ \hline
SimCLR$^*$ \cite{simclr_2020}   & 85.30          & 56.50         & 78.34        & 37.16          \\
BYOL$^*$ \cite{byol_2020}       & 87.54          & 57.24         & 85.62        & 37.65          \\ 
SCE$^*$ \cite{sce_2023}         & 88.54          & {\ul59.97}    & 85.09        & 40.48         \\
MoCoV2$^*$ \cite{mocov2_2020}   & 87.82          & 57.29         & 84.66        & 37.77          \\
UnMix$^*$ \cite{unmix_2022}     & 87.99          & 59.11         & 85.05        & 38.65          \\ \hline
NNCLR$^*$ \cite{nnclr_2021}     & 85.19          & 50.54         & 81.31        & 30.93          \\
SNCLR$^*$ \cite{snclr_2023}     & 87.36          & 58.65         & {\ul86.02}        & {\ul41.92}          \\
CMSF$^*$ \cite{cmsf_2022}       & {\ul89.30}     & 55.57         & 84.11        & 36.79          \\
MSF$^*$ \cite{msf_2021}         & 88.24          & 52.32         & 84.09        & 35.29          \\ 
\rowcolor[HTML]{ECF4FF} 
MNN                      & \textbf{89.81} & \textbf{61.96} & \textbf{86.65}       & \textbf{42.28} \\ \hline
\end{tabular}
\label{table: knn_main}
\end{table}

\subsubsection{Analysis of experimental results and insights}
Given that MNN solely modifies the method of weight adjustment and introduces a mixture operation compared to MSF, our analysis reveals some intriguing patterns regarding the performance gains of MNN over MSF across various evaluation metrics: \textit{(i) Larger datasets yield enhanced performance}. When the number of dataset classes remains constant, the training set of STL-10, which is nearly twice the size of CIFAR-10's training set, showcases more substantial performance improvements for MNN compared to MSF. Specifically, MNN's performance on STL-10 saw improvements of 3.56\% and 2.56\% in two separate experiments, whereas the improvements on CIFAR-10 were 1.53\% and 1.57\%, respectively. \textit{(ii) Greater class diversity correlates with superior performance gains}. With dataset size constant, CIFAR-100's tenfold increase in class diversity over CIFAR-10 leads to a notable performance decline for MSF, likely due to the higher probability of incorporating false neighbors with an increased number of classes. In contrast, MNN thrives in this scenario by effectively countering the effects of false neighbors, enhancing the model's performance.

In summary, our experimental results illustrate that MNN achieves heightened performance gains with larger datasets and adeptly mitigates the influence of false neighbors in scenarios with an increased number of classes.

\subsection{Ablation studies}\label{43}
In this section, we will explore the components and parameters of MNN, such as data augmentation, the number of neighbors (K), the size of the support set, and the neighbor selection strategy. By default, MNN utilizes '(s/w)' data augmentation and sets the number of neighbors (K) to 5, employing the top-K nearest-neighbor selection strategy unless stated otherwise. To assess the purity of the neighbor set, we define it as the ratio of neighbor samples with the same label as the positive sample to the total number of samples. \textbf{\textit{Notably, since STL-10 is primarily used for unsupervised pre-training, its training samples containing labels make up approximately 5\% (5,000 out of 105,000) of the overall training dataset size. Consequently, to utilize the labels for precise purity analysis, we conduct purity-related experiments on the remaining three datasets}}.

\begin{figure}[t]
	\centering
	\subfigure[]{
		\begin{minipage}[t]{0.35\linewidth}
			\centering
			\includegraphics[width=\linewidth]{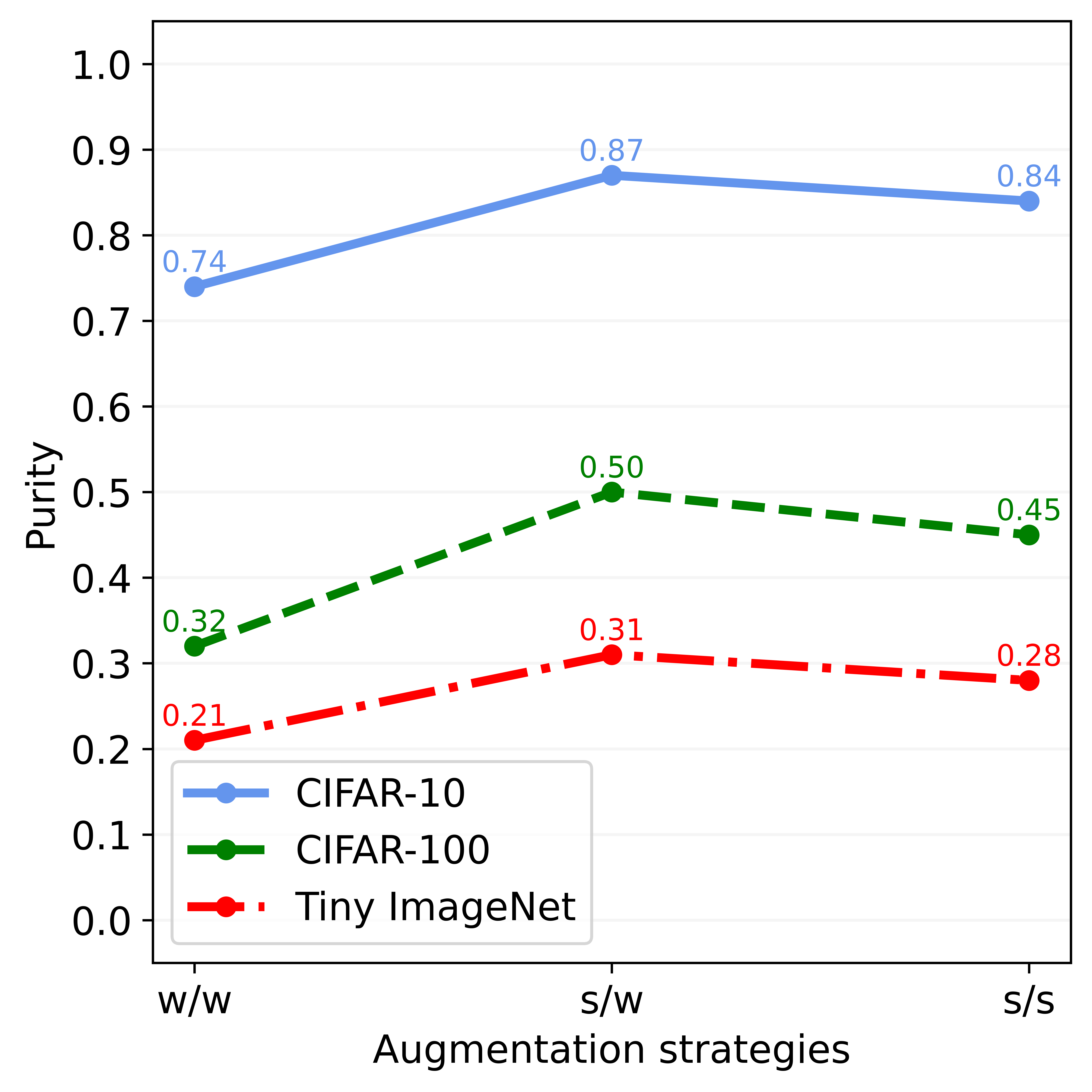}
		\end{minipage}
   \label{figure: ablation_purity_aug}
	}%
	\subfigure[]{
		\begin{minipage}[t]{0.35\linewidth}
			\centering
			\includegraphics[width=\linewidth]{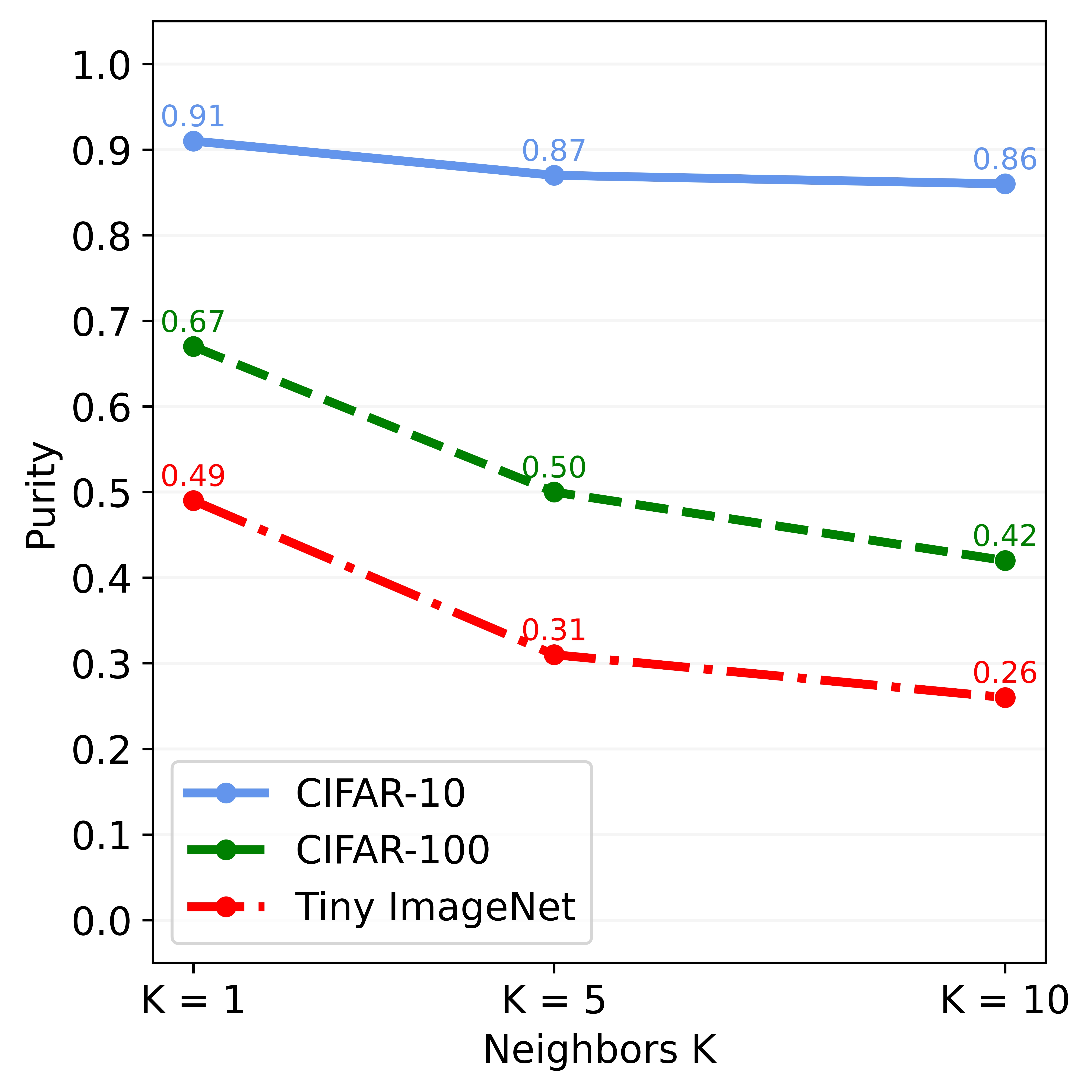}
		\end{minipage}
  \label{figure: ablation_purity_ks}
	}%
 
	\subfigure[]{
		\begin{minipage}[t]{0.35\linewidth}
			\centering
			\includegraphics[width=\linewidth]{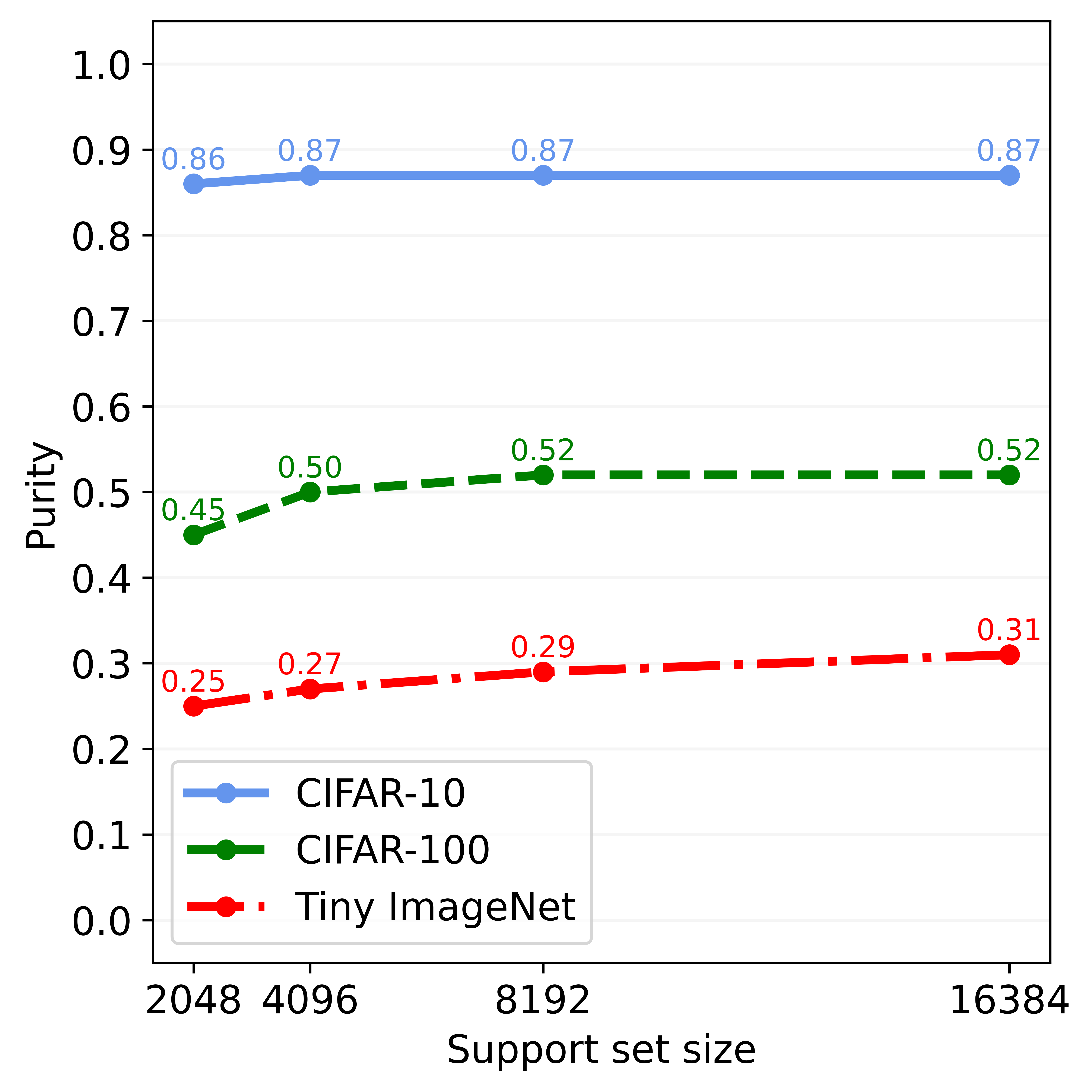}
		\end{minipage}
  \label{figure: ablation_purity_set_size}
	}%
    \subfigure[]{
		\begin{minipage}[t]{0.35\linewidth}
			\centering
			\includegraphics[width=\linewidth]{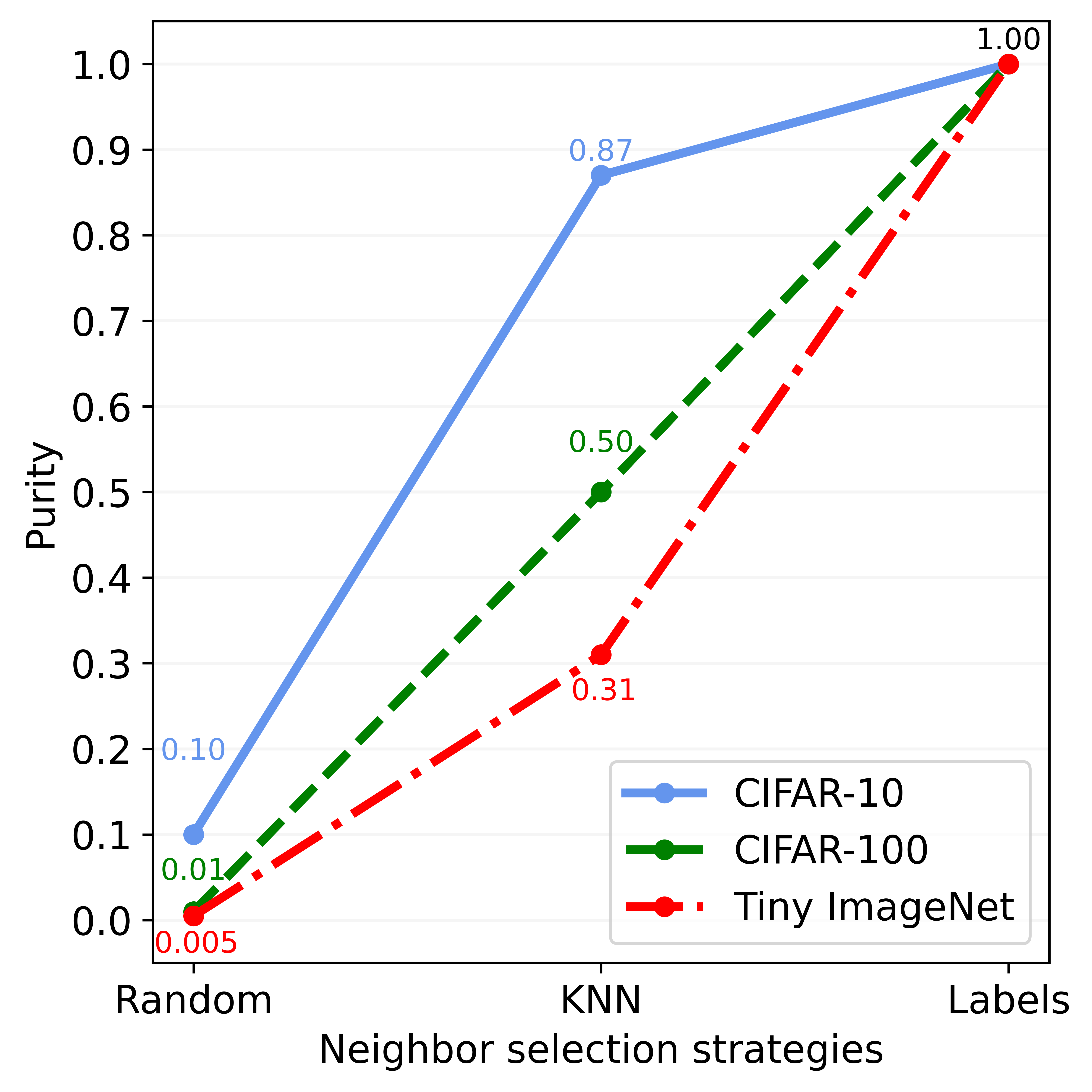}
		\end{minipage}
   \label{figure: ablation_purity_nn_selection}
	}%
	\centering
	\caption{The effect of different component variations on purity. (a) The effect of data augmentation on purity. (b) The effect of the number of nearest-neighbors on purity. (c) The effect of support set size on purity. (d) The effect of different nearest-neighbor selection strategies on purity. Purity represents the ratio of neighbors with the same label as the positive sample to the total number of samples.}
     \label{figure: ablation_purity}
\end{figure}

\subsubsection{Data augmentation strategies}\label{431}
Data augmentation is a key factor in promoting contrastive self-supervised learning by encouraging the development of semantic features that remain consistent under geometric transformations. However, aggressive data augmentation, as in '(s/s),' may hinder the ability of the teacher network to identify suitable neighbor samples for $z^2$. Our hypothesis is confirmed by Fig.~\ref{figure: ablation_purity_aug}, which illustrates that strong data augmentation reduces the purity of the neighbor set, introducing more false neighbor samples. In Table~\ref{table: ablation_aug}, we present linear evaluation results for various data augmentation strategies. In the '(w/w)' configuration, where MNN lacks strong data augmentation, it might face challenges in learning semantic features. Nevertheless, MNN maintains high performance compared to MSF, indicating that the mixture operation could potentially function as a form of strong data augmentation in domains with limited prior knowledge.

\begin{table}[h!]
\centering
\setlength{\belowcaptionskip}{6 pt}
\caption{Comparison of data augmentation strategies. We present the linear evaluation accuracy of MNN and MSF using various data augmentation strategies. Both MSF and MNN attained their highest accuracy when strong data augmentation was applied to the student network and weak data augmentation to the teacher network. This outcome aligns with our expectation that aggressive data augmentation can impact the accuracy of neighbor selection for $z^2$. $^*$ denotes the results obtained by different ways of data augmentation based on the official code.}
\begin{tabular}{lcccc}
\hline
Method   & CIFAR-10       & CIFAR-100      & STL-10         & Tiny ImageNet  \\ \hline
MSF$^*$ (w/w) & 75.54          & 30.06          & 77.60          & 18.55          \\
\rowcolor[HTML]{ECF4FF} 
MNN (w/w) & \textbf{84.32} & \textbf{50.71} & \textbf{82.08}          & \textbf{34.85} \\ \hline
MSF$^*$      & 89.94          & 59.94          & 88.05          & 42.68          \\
\rowcolor[HTML]{ECF4FF} 
MNN      & \textbf{91.47} & \textbf{67.56} & \textbf{91.61}          & \textbf{50.70} \\ \hline
MSF$^*$ (s/s) & 89.73          & 58.00          & 87.18              & 40.15          \\
\rowcolor[HTML]{ECF4FF} 
MNN (s/s) & \textbf{90.65} & \textbf{64.91} & \textbf{91.15}      & \textbf{48.07} \\ \hline
\end{tabular}
\label{table: ablation_aug}
\end{table}
\subsubsection{Number of nearest-neighbors}\label{432}
To increase the semantic richness of positive samples, we consider the nearest neighbors retrieved by the teacher network as additional positives. Generally, performance tends to improve with increasing K. Fig.~\ref{figure: ablation_purity_ks} illustrates how the purity of MNN changes with different K values. As K increases, there is a higher risk of introducing false neighbor samples, resulting in decreased purity. However, setting K to 1 yields a purer neighbor set but may limit semantic diversity. Table~\ref{table: ks_ablation} presents the linear classification accuracies at different K values. We observed that for MSF and SNCLR, the optimal K value for peak performance was $K = 1$. In contrast, for the majority of datasets, MNN exhibited its optimal performance when K was set to 5. This suggests that MSF and SNCLR are more susceptible to false neighbors, while MNN effectively mitigates this issue, striking a balance between neighbor set purity and diversity. Therefore, MNN successfully integrates nearest neighbors into SSL, enhancing the diversity of semantic features. Additionally, when $K = 1$, MSF resembles a variant of MNN that does not utilize the mixture operation.

\begin{table}[htb]
\centering
\setlength{\belowcaptionskip}{6 pt}
\caption{Analysis of the number of neighbors (K). $^*$ denotes the result obtained by adjusting the number of nearest neighbors according to the official code.}
\begin{tabular}{lccc}
\hline
        & $K=1$ & $K=5$           & $K=10$ \\ \hline
        &          & CIFAR-10           &      \\ \hline
MSF$^*$ \cite{msf_2021}     & \textbf{91.55}    & 89.94              & 89.46   \\
SNCLR$^*$ \cite{snclr_2023}  & \textbf{90.04}    & 88.86              & 89.81    \\
\rowcolor[HTML]{ECF4FF} 
MNN     & 91.41    & \textbf{91.47}                & 91.24     \\ \hline
        &     & CIFAR-100     &      \\ \hline
MSF$^*$ \cite{msf_2021}     & \textbf{65.22}    & 59.94              & 57.25     \\
SNCLR$^*$ \cite{snclr_2023}  & \textbf{65.21}    & 65.19              & 65.03     \\
\rowcolor[HTML]{ECF4FF} 
MNN     & 66.98    & \textbf{67.56}              & 67.13     \\ \hline
        &     & STL-10     &      \\ \hline
MSF$^*$ \cite{msf_2021}     & \textbf{90.28}    & 88.05              & 86.99     \\
SNCLR$^*$ \cite{snclr_2023}  & 90.84             & \textbf{90.93}              & 90.26     \\
\rowcolor[HTML]{ECF4FF} 
MNN     & \textbf{91.69}    & 91.61              &  91.59     \\ \hline
        &           & Tiny ImageNet   &      \\ \hline
MSF$^*$ \cite{msf_2021}     & \textbf{47.06}     & 42.68           &  39.68    \\
SNCLR$^*$ \cite{snclr_2023}  & 50.04     & \textbf{50.15}           &   49.30   \\
\rowcolor[HTML]{ECF4FF} 
MNN     & 49.70  & \textbf{50.70}     & 49.62  \\ \hline
\end{tabular}
\label{table: ks_ablation}
\end{table}
\subsubsection{Size of the support set}\label{433}
The size of the support set has the potential to influence its ability to accurately reflect the data distribution. As depicted in Fig.~\ref{figure: ablation_purity_set_size}, increasing the support set size leads to higher neighbor set purity, indicating that the current instance can find suitable nearest neighbors more effectively. However, the model classification performance is shown in Table~\ref{table: support_set_size_ablation}, and we hypothesize that an excessively large support set may hinder timely updates of sample features, consequently impacting the performance of the model. Therefore, we strike a balance between aligning with the data distribution and ensuring that sample features are updated promptly.

\begin{table}[htb]
\centering
\setlength{\belowcaptionskip}{6 pt}
\caption{Analysis of support set size.}
\begin{tabular}{lcccc}
\hline
Size      & CIFAR-10       & CIFAR-100      & STL-10     & Tiny ImageNet  \\ \hline
2048  & 91.32          & 66.68              & 91.55            & 49.68           \\
4096  & \textbf{91.47} & \textbf{67.56}    & 91.50                 & 49.80          \\
8192  & 91.31    & 67.33                  & 91.36                  & 49.93    \\
16384 & 91.27          & 67.24           & \textbf{91.61}                       & \textbf{50.70} \\ \hline
\end{tabular}
\label{table: support_set_size_ablation}
\end{table}
\subsubsection{Neighbors selection strategies}\label{434}
We experimented with two approaches for selecting neighbor samples, aside from using cosine similarity. In the random selection method, K samples are randomly chosen from the support set to serve as neighbor samples for the current sample $z^2$. This approach is akin to UnMix but without the need to mix images in the pixel space and perform extra forward propagation.  Consequently, our method incurs minimal computational overhead. To explore the potential of MNN, we used an Oracle algorithm to select samples with the same label as $z^2$ as its corresponding neighbors in the support set. To achieve this, we preserved the labels of these elements while updating the elements of the support set. This approach can be considered as a form of supervised learning without label predictions, achieving a neighbor set with 100\% purity. Figure~\ref{figure: ablation_purity_nn_selection} illustrates the purity of the neighbor set under various settings, while Table~\ref{table: nn_select_ablation} details the linear evaluation accuracy within these scenarios. These data demonstrate the significant performance and efficiency gains achieved by MNN. These outcomes highlight the critical role of addressing the false neighbor issue to optimize performance.

\begin{table}[htb]
\centering
\setlength{\belowcaptionskip}{6 pt}
\caption{Comparison of neighbor selection strategies. STL-10 is mainly employed for unsupervised pre-training, with labeled training samples constituting only around 5\% of the total training dataset. Consequently, we focus our evaluation on the remaining three datasets in their labeled configurations. The $^*$ symbol indicates that these results were obtained through the reproduction of the official code.}
\begin{tabular}{lcccc}
\hline
Method     & CIFAR-10             & CIFAR-100     & STL-10       & Tiny ImageNet        \\ \hline
UnMix$^*$ \cite{unmix_2022}     & \textbf{90.37}       & 65.30         & 90.51                 & 47.29       \\
\rowcolor[HTML]{ECF4FF} 
MNN(Random)        & 90.25          & \textbf{65.88}   & \textbf{90.60}             & \textbf{47.50}         \\ \hline
MSF$^*$ \cite{msf_2021}       & 89.94         & 59.94                & 88.05                   & 42.68          \\
\rowcolor[HTML]{ECF4FF} 
MNN        & \textbf{91.47} & \textbf{67.56}    & \textbf{91.61}                  & \textbf{50.70} \\ \hline
Supervised \cite{ressl_2021} & \textbf{94.22}                & 74.66            & 82.55            & 59.26                \\
\rowcolor[HTML]{ECF4FF} 
MNN(Labels)        & 94.00          & \textbf{74.84}         & -         & \textbf{60.33}               \\ \hline
\end{tabular}
\label{table: nn_select_ablation}
\end{table}
\subsection{Analysis and discussion}\label{44}
We will conduct a detailed analysis of the impact of weight adjustments, which are designed to differentiate between positive and neighbor samples. Following this, we will explore the influence of mixing positive and neighbor samples on the model, elucidating how the mixture mechanism of MNN enhances the optimization between $z^2$ and $p^1$. These analyses will gradually unveil the various factors at play within the MNN model. Finally, we will apply the proposed WSE and mixture operation with other methods to verify the generalization of the proposed approach.

\begin{figure}[t]
	\centering
	\subfigure[Inconsistency]{
		\begin{minipage}[t]{0.32\textwidth}
			\centering
			\includegraphics[width=\textwidth]{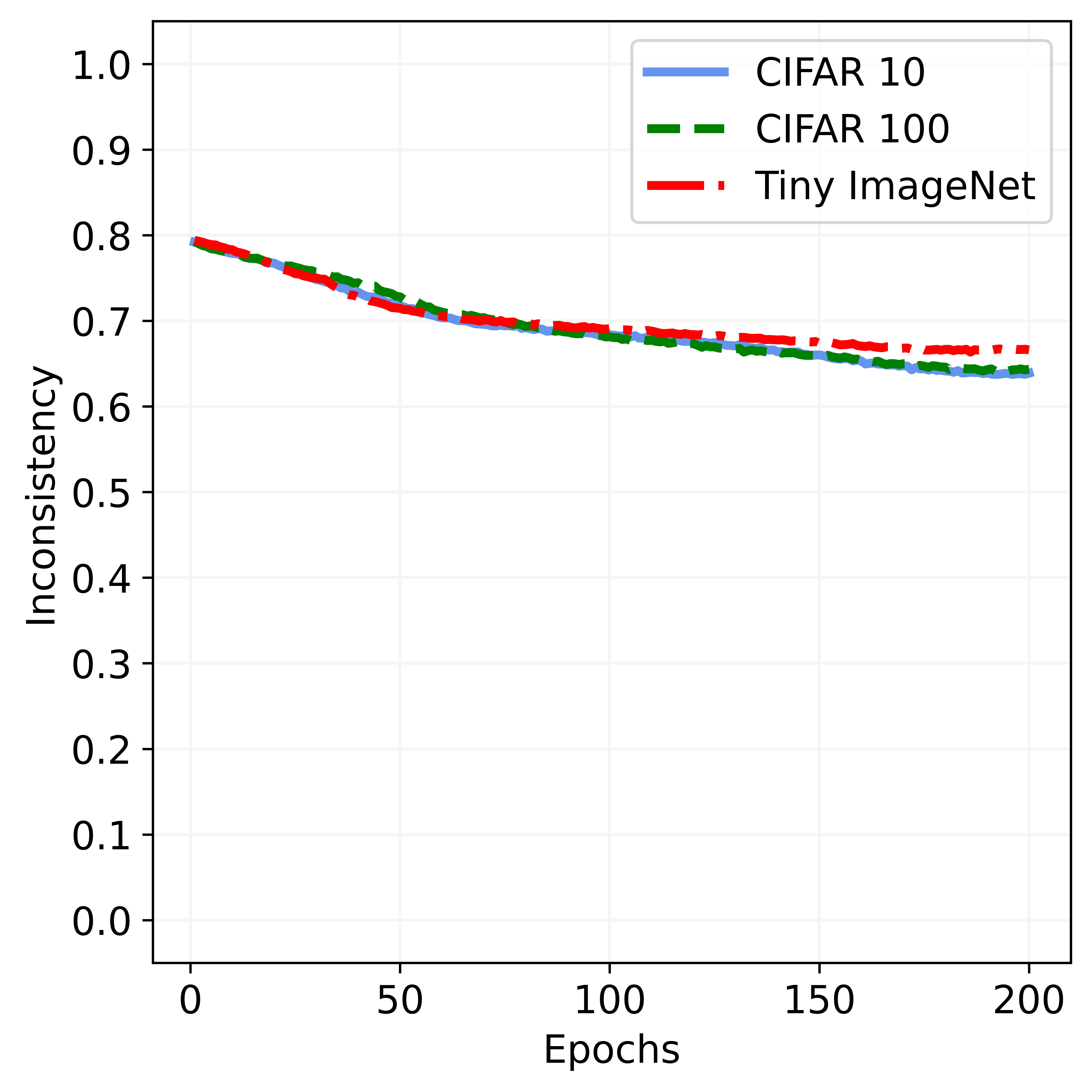}
		\end{minipage}
  \label{figure: inconsistency_smsf}
	}%
	\subfigure[Entropy]{
		\begin{minipage}[t]{0.32\textwidth}
			\centering
			\includegraphics[width=\textwidth]{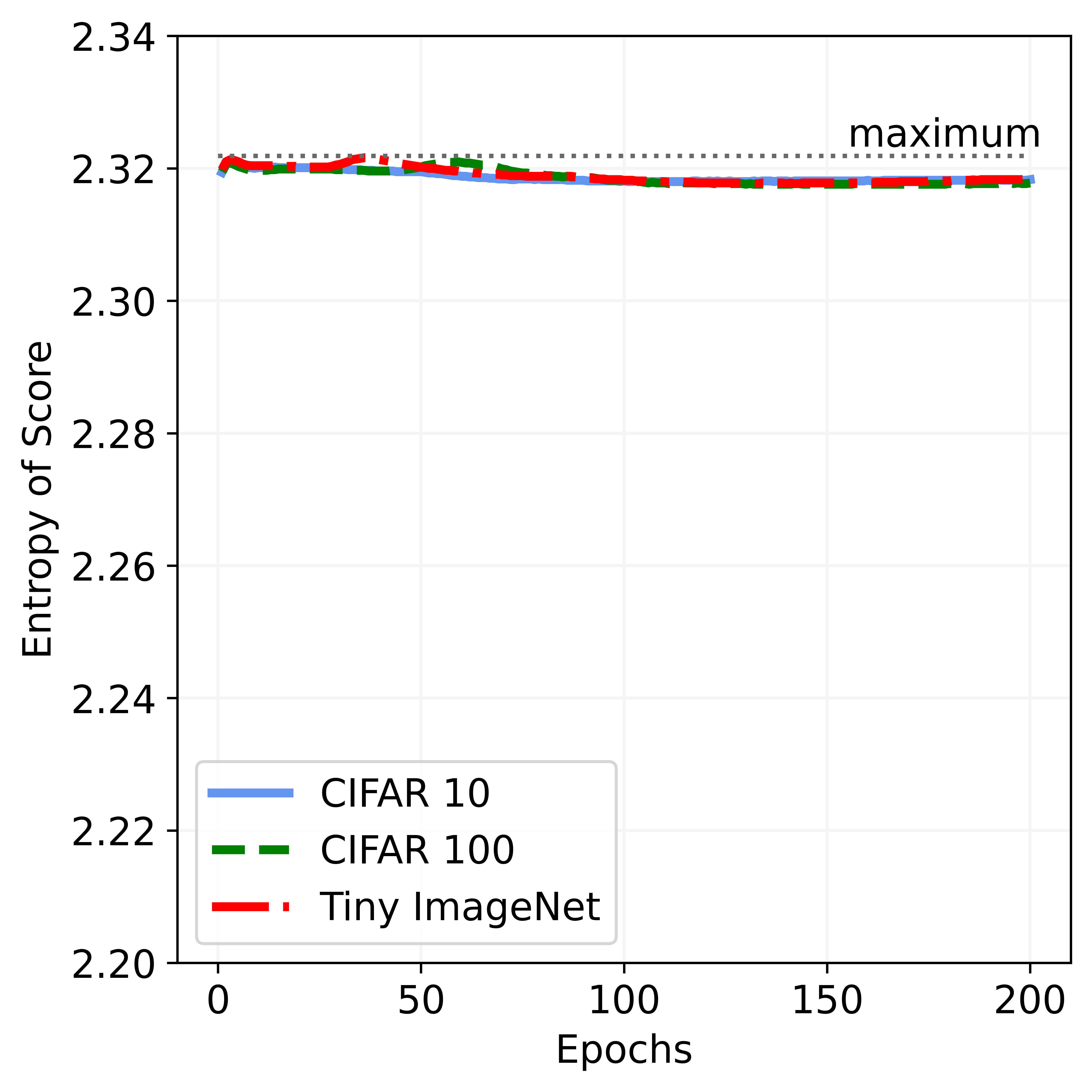}
		\end{minipage}
  \label{figure: entropy_smsf}
	}%
    \subfigure[Purity]{
		\begin{minipage}[t]{0.32\textwidth}
			\centering
			\includegraphics[width=\textwidth]{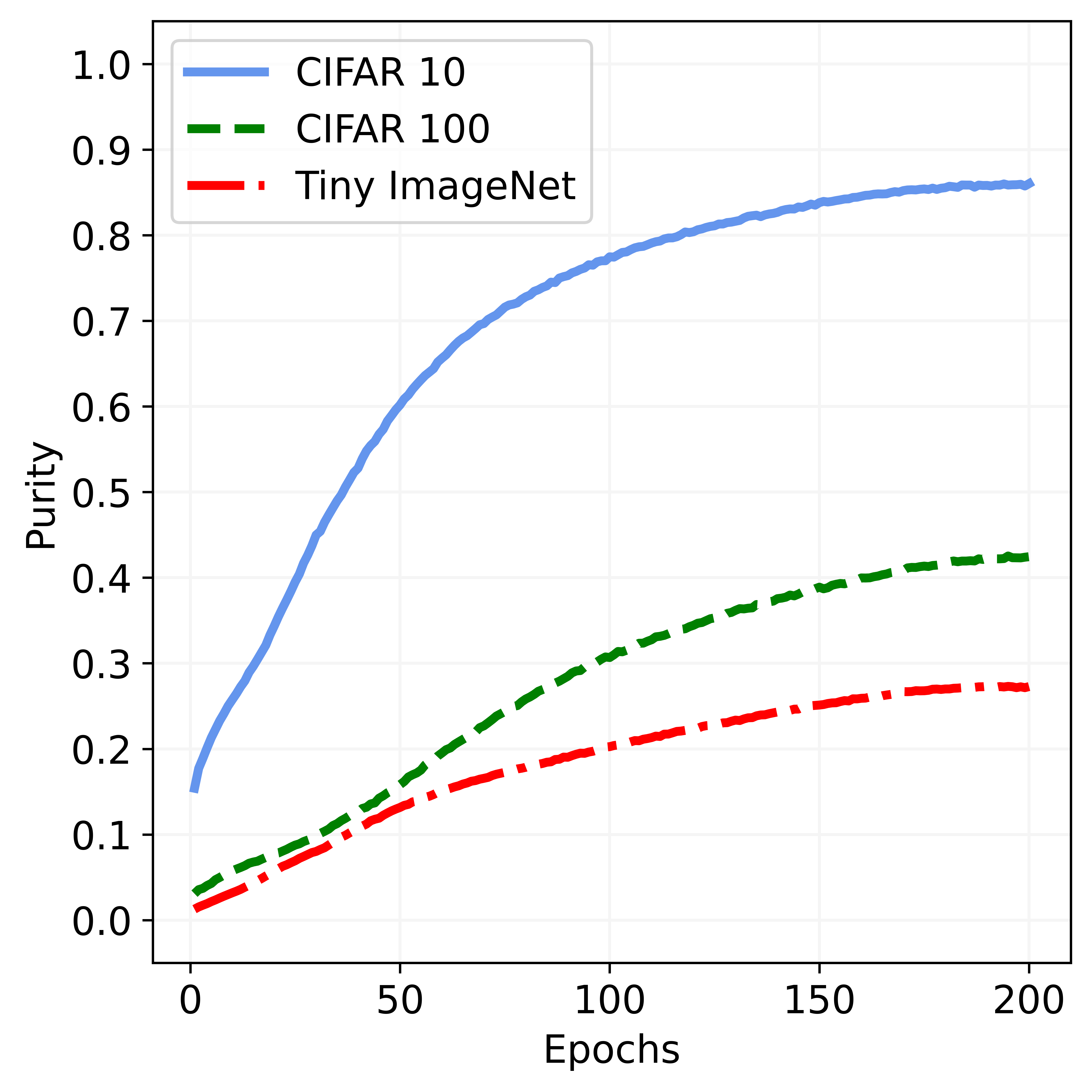}
		\end{minipage}
    \label{figure: purity_smsf}
	}%
	\centering
	\caption{Details of using the Cross-Attention Score in the MSF approach. Fig.~(a) illustrates the inconsistency across different datasets. Fig.~(b) presents the entropy of the CAS distribution across different datasets. Fig.~(c) shows the variation in purity across different datasets. Although purity varies significantly across datasets, it is noteworthy that the entropy of the CAS distribution remains relatively high. This suggests that CAS may assign higher weights to false neighbors, thereby confusing the model in task optimization.}
     \label{figure: purity_smsf_entropy}
\end{figure}
\subsubsection{Diverse strategies for weight adjustment}\label{441}
By assigning appropriate weights to different subtasks within the loss function, we can adapt the focus of the model to better fit the parameters. In this context, we aim to substitute the weight adjustment methods of MNN and analyze how different weight adjustment techniques affect the model. To achieve this, we perform experiments using the Mean Squared Error from MSF \cite{msf_2021} and the Cross-Attention Score, which comprises identity mappings without parameters, as employed in SNCLR \cite{snclr_2023}:

\begin{subequations}
  \begin{align}
    \label{eq:msf}
w_i^{MSE} & = \frac{1}{K+1} \ ,\\
    \label{eq:attention}
w_i^{CAS} &= \begin{cases}
 1 , &\ i= 0 \\
\frac{1}{{\textstyle \gamma_i}}\frac{ {\textstyle exp(cos(z^2_i, q^1))}}{ {\textstyle \sum_{k=1}^{K} exp(cos(z^2_k, q^1))}}, &\mbox{ otherwise } \ ,
\end{cases}
\end{align}
\end{subequations}
where $q^1$ is the positive embedding from the student network, and $\gamma_i$ represents the scaling factor. If we rearrange the ordered set of neighbors ${\{z^2_i\}}_{i=1}^K$ using the values computed by the CAS, we obtain a different ordered set, ${\{z^{2'}_j\}}_{j=1}^K$. To quantify the inconsistency in the order of neighbors, we define it as follows:

\begin{equation}
\mbox{Inconsistency} = \frac{{\textstyle \sum_{k=1}^{K}\mathds{1}_{[z^2_k \ne z^{2'}_k]} }}{K} \ ,
\label{eq:inconsistency}
\end{equation}
where $\mathds{1}_{[z^2_k \ne z^{2'}_k]} \in {\{0, 1\}}$ is an indicator function that equals 1 when $z^2_k$ is not equal to $z^{2'}_k$.

As shown in Fig.~\ref{figure: inconsistency_smsf}, the observed inconsistency appears to be a common occurrence across various datasets. This inconsistency may result from differences in data augmentation between the student and teacher networks, as well as the presence of false neighbors. Although it may shuffle neighbor samples that are semantically similar to the positive samples to the front of the ordered set (as shown in \ref{A3}), our analysis, presented in Fig.~\ref{figure: entropy_smsf}, consistently reveals that the Cross Attention Score results in high levels of entropy in the weight distribution ${\{w^{CAS}_i\}}_{i=1}^K$. Additionally, in conjunction with neighbor sample purity (Fig.~\ref{figure: purity_smsf}), it is evident that this high entropy level signifies the inaccuracy of the CAS, as larger weights are assigned to False Nearest-Neighbors. The lower section of Table~\ref{table: mnn_important_component}, excluding the mixture component, corroborates our observation: the weight adjustment approach in WSE markedly improves model performance compared to CAS.

\begin{table}[h]
\centering
\setlength{\belowcaptionskip}{6 pt}
\captionsetup{position=above} 
\caption{Analysis of crucial components for MNN. We systematically investigate the effect of different components in MNN on the accuracy of linear evaluation. The \colorbox[HTML]{ECF4FF}{highlights} indicate changes compared to the default MNN settings. The lower part shows the superiority of our WSE over the CAS. In the upper part, we can clearly observe that the mixture operation significantly improves the accuracy of linear evaluation across all weight adjustment approaches. The $^*$ symbol indicates that these results were obtained through the reproduction of the official code.}
\resizebox{\linewidth}{!}
{
\begin{tabular}{l|ccc|c|cccc}
\hline
Method & MSE      & CAS  & WSE  & Mixture  & CIFAR-10  & CIFAR-100  & STL-10 & Tiny ImageNet \\ \hline
MNN    & \ding{55}  &\ding{55} & \ding{51}  & \ding{51} & 91.47    & 67.56   & 91.61  & 50.70   \\
       & \ding{55}  & \colorbox[HTML]{ECF4FF}{\ding{51}} & \ding{55}  & \ding{51}     & 91.33         & 67.32     & 91.20   & 50.06        \\
        & \colorbox[HTML]{ECF4FF}{\ding{51}} & \ding{55}   &\ding{55}      & \ding{51}  & 91.20          & 65.16    & 90.08       & 47.16         \\
       &\ding{55}   &\ding{55} & \ding{51}  & \colorbox[HTML]{ECF4FF}{\ding{55}} & 91.40    & 65.55  & 91.20   & 49.27   \\
       & \ding{55}  & \colorbox[HTML]{ECF4FF}{\ding{51}} & \ding{55}  & \colorbox[HTML]{ECF4FF}{\ding{55}} & 90.91    & 63.48  & 91.13   & 47.32   \\  \hline
MSF$^*$ \cite{msf_2021}   & \colorbox[HTML]{ECF4FF}{\ding{51}}  & \ding{55}   & \ding{55}     & \colorbox[HTML]{ECF4FF}{\ding{55}}
& 89.94    & 59.94  & 88.05   & 42.68   \\ \hline
\end{tabular}
}
\resizebox{\linewidth}{!}
{
\par  
MSE: Mean Squared Error, CAS: Cross-Attention Score, WSE: Weighted Squared Error
}

\label{table: mnn_important_component}
\end{table}

\subsubsection{The meaning of mixture}\label{442}
The primary goal of incorporating the mixture operation in MNN is to mitigate the noise introduced by false neighbors. As is evident from the upper section of Table~\ref{table: mnn_important_component}, the inclusion of the mixture component significantly enhances accuracy across various weight adjustment methods. \ref{A2} illustrates that the mixture operation in MNN refines the optimization process between $z^2$ and $p^1$, enhancing the performance of the model with neighbor samples by considering them as additional positive samples. This analysis provides valuable insights into the role of the mixture operation within MNN.

\subsubsection{Approach generality}\label{443}
We apply Weighted Squared Error and mixture operations to other contrastive self-supervised methods, namely CMSF and SNCLR. In SNCLR, we incorporate weak data augmentation to maintain the semantics of the output embedding in the teacher network branch. Notably, this adaptation consistently improves the performance of all methods on most benchmark image datasets, as illustrated in Table~\ref{table: approach_generality}. Furthermore, we note that the improvement in SNCLR is relatively modest. We hypothesize that this may be attributed to the fact that SNCLR uses InfoNCE loss \cite{infonce_2018} thereby introducing noise disturbance from false negative samples \cite{false_negative_samples_2019}.

\begin{figure}[t]
	\centering
	\subfigure[MoCoV2]{
		\begin{minipage}[t]{0.4\textwidth}
			\centering
			\includegraphics[width=\textwidth]{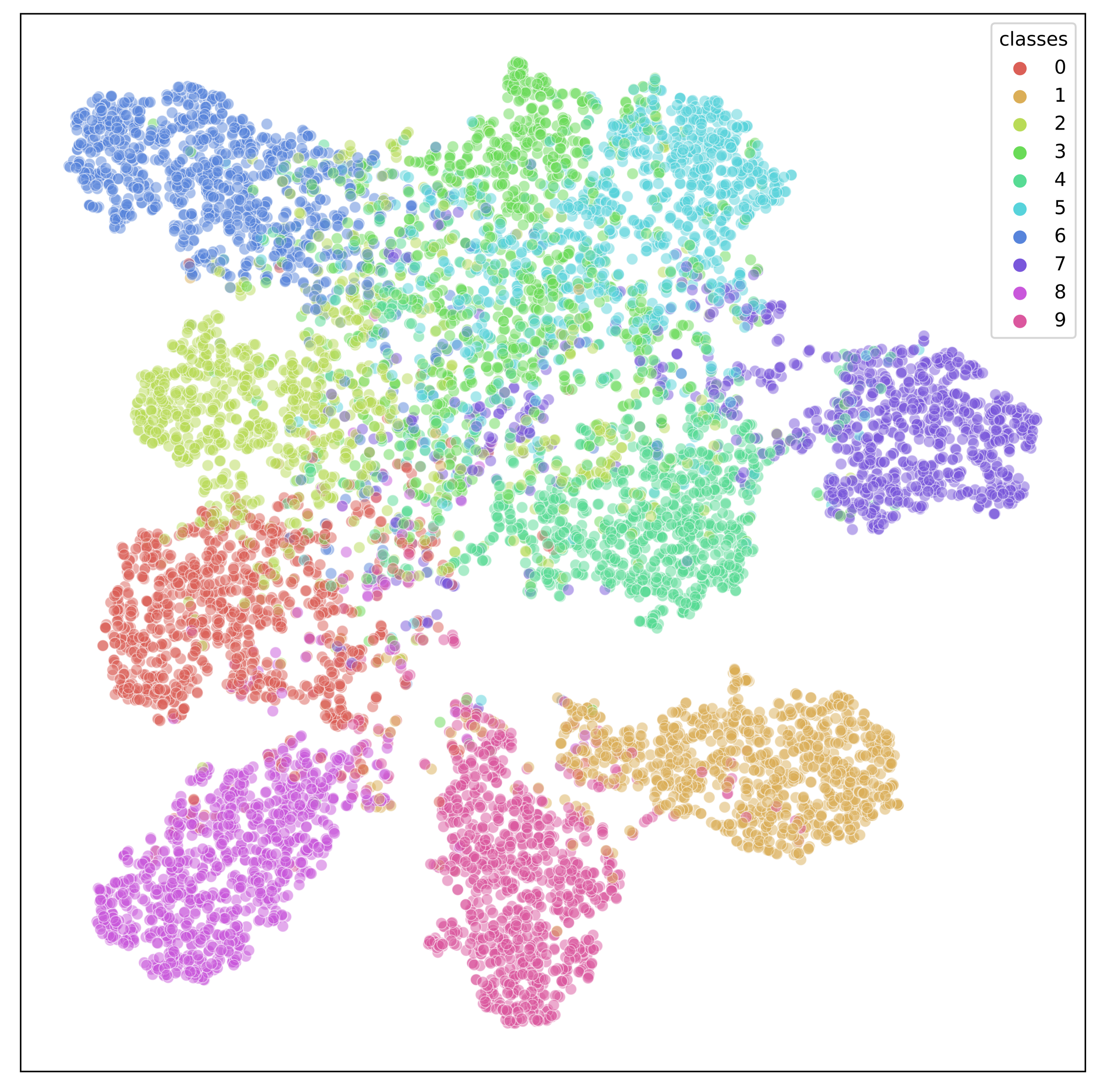}
		\end{minipage}
  \label{figure: illustration_tsne_msf}
	}%
	\subfigure[MNN]{
		\begin{minipage}[t]{0.4\textwidth}
			\centering
			\includegraphics[width=\textwidth]{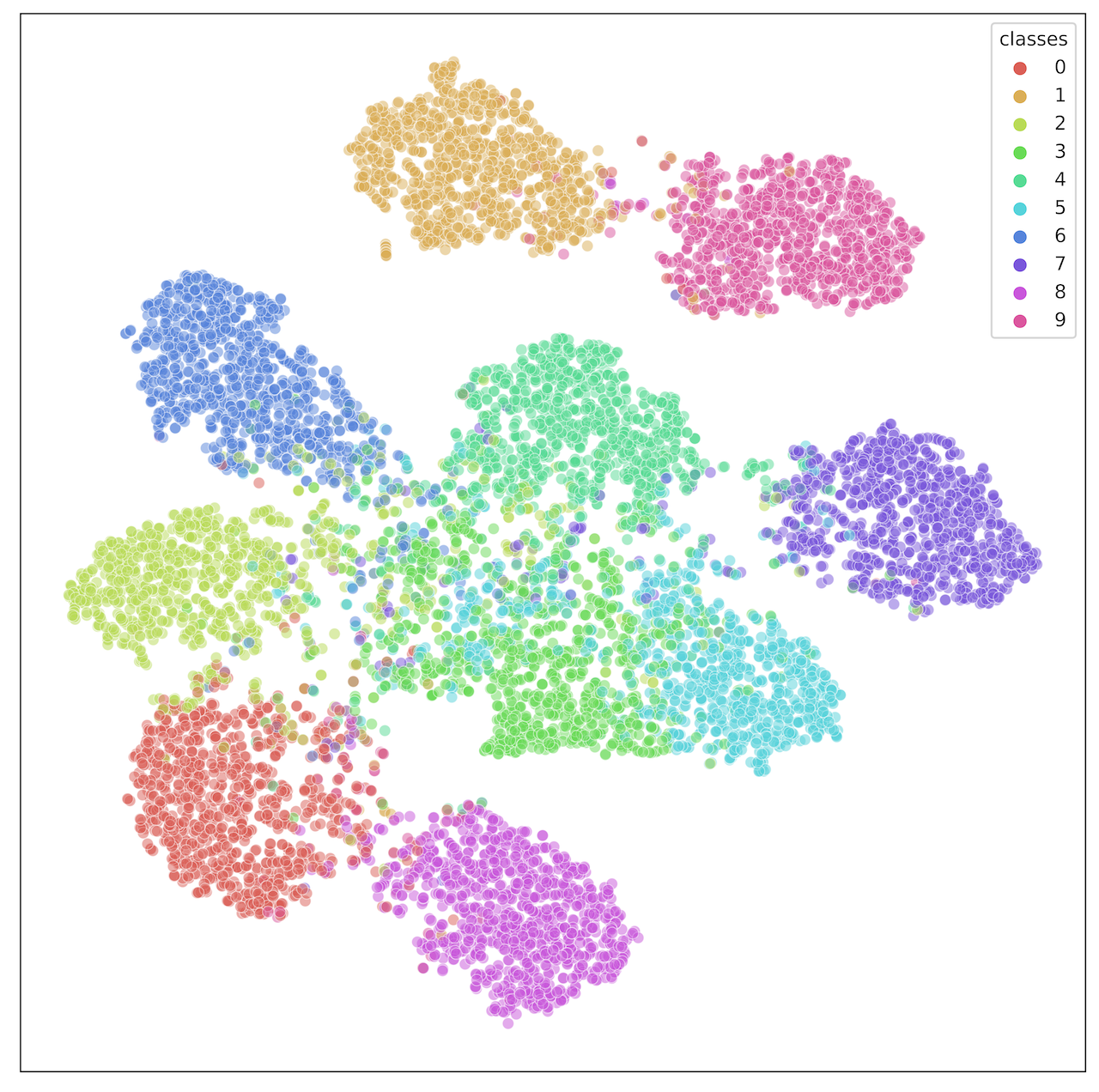}
		\end{minipage}
  \label{figure: illustration_tsne_mnn}
	}%
	\centering
	\caption{Visualization of t-SNE features for MoCoV2 and MNN on CIFAR-10.}
     \label{figure: illustration_tsne}
\end{figure}

\begin{table}[htb]
\centering
\setlength{\belowcaptionskip}{6 pt}
\caption{Generalization of the approach. The upper section of the table displays the linear evaluation accuracy, while the lower section corresponds to the K-nearest neighbor classification with $K = 200$. Notably, MSF and CMSF employ MSE, whereas SNCLR uses CAS. The $^*$ symbol indicates that these results were obtained through the reproduction of the official code.}
\resizebox{\linewidth}{!}
{
\begin{tabular}{lp{0.0001\textwidth}ccccp{0.0001\textwidth}cccc}
\hline
     &  & \multicolumn{4}{c}{MSE or CAS}        &  & \multicolumn{4}{c}{WSE and Mixture (Ours)} \\ \cline{3-6} \cline{8-11}
     &  & CIFAR-10 & CIFAR-100 &STL-10  & Tiny ImageNet &  & CIFAR-10   & CIFAR-100  &STL-10  & Tiny ImageNet  \\ \hline
MSF$^*$ \cite{msf_2021} &  & 89.94    & 59.94  & 88.05   & 42.68   &  & 91.47      & 67.56   & 91.61    & 50.70    \\
CMSF$^*$ \cite{cmsf_2022} &  & 91.00    & 62.37  & 88.21   & 44.50   &  & 91.81      & 67.72   & 91.80    & 50.73    \\ 
SNCLR$^*$ \cite{snclr_2023} &  & 88.86    & 65.19 & 90.93    & 50.15   &  & 90.48      & 67.67  & 91.25     & 51.02    \\ 
\hline\hline
MSF$^*$ \cite{msf_2021} &  & 88.24    & 52.32  & 84.09   & 35.29   &  & 89.81      & 61.96   & 86.65   & 42.28    \\
CMSF$^*$ \cite{cmsf_2022} &  & 89.30    & 55.57  & 84.11   & 36.79   &  & 90.08      & 61.05   & 87.59    & 42.83    \\ 
SNCLR$^*$ \cite{snclr_2023} &  & 87.36    & 58.65 & 86.02    & 41.92   &  & 87.96      & 60.50  & 85.40     & 42.48    \\ \hline
\end{tabular}
}
\label{table: approach_generality}
\end{table}
\subsection{Visualization of features}\label{45}
We also demonstrated the semantic features acquired by our proposed method, MNN, by performing t-SNE \cite{t-SNE_2008} visualization on the CIFAR-10 dataset. The visualization results in Fig.~\ref{figure: illustration_tsne} reveal that MNN exhibits distinct category boundaries while maintaining compact embeddings within categories.

\section{Conclusion}\label{5}
In this work, we introduced MNN, a simple self-supervised visual representation learning framework that enhances the diversity of semantic features learned by the model. MNN employs a straightforward loss function and incorporates a mixture operation to efficiently include nearest neighbors in instance discrimination tasks. Our experiments highlight the ability of the MNN to effectively reduce the impact of false neighbors on the model with minimal computational overhead. In the future, we plan to extend the concept of mixture operations to various unsupervised learning scenarios, particularly those involving imbalanced or fine-grained datasets.

\section*{Acknowledgement} \label{6}
This work was supported by the National Natural Science Foundation of China under Grant No. 61906098.
\bibliographystyle{elsarticle-num} 
\biboptions{sort&compress}
\bibliography{ref} 

\appendix
\onecolumn
\section{Pseudo-code}\label{A1}
In this section, we provide the PyTorch-style pseudo-code for MNN (as shown in Algorithm~\ref{algo:algo-mnn}). Our code is publicly available at \href{https://github.com/pc-cp/MNN}{https://github.com/pc-cp/MNN}. Notably, in comparison to MSF, MNN introduces only minimal computational overhead, as \colorbox[HTML]{ECF4FF}{highlighted} in the pseudo-code. However, this minor adjustment results in substantial performance enhancements.
\definecolor{commentcolor}{RGB}{110,154,155}   
\newcommand{\PyComment}[1]{\ttfamily\textcolor{commentcolor}{\# #1}}  
\newcommand{\PyCode}[1]{\ttfamily\textcolor{black}{#1}} 
\newcommand{\PyNew}[1]{\ttfamily\textcolor{red}{#1}}  
\begin{center}  
\begin{minipage}{0.94\textwidth}
\begin{algorithm}[H]
\footnotesize 
\begin{spacing}{1.0}
\SetAlgoLined
    \PyComment{$\mathtt {F_s}$, $\mathtt {F_t}$: encoder for student, teacher, $\mathtt {F\_\triangleq g\_(f\_(\cdot))}$} \\
    \PyComment{$\mathtt {h_s}$: predictor for student}\\
    \PyComment{$\mathcal{S}$: support set(CxQ)} \\
    \PyComment{m: momentum for teacher} \\
    \PyComment{topk: number of nearest-neighbors} \\
    \vspace*{\baselineskip}
    \PyCode{$\mathtt {F_t}$.params = $\mathtt {F_s}$.params} \ \ \ \ \ \ \PyComment{initialize $\mathtt {teacher}$} \\
    \PyComment{load a minibatch X with N samples}\\
    \PyCode{for X in loader: } \\
    \Indp   
        \PyComment{random augmentation} \\
        \PyCode{$\mathtt {X_s}$, $\mathtt {X_t}$ = strong\_aug(X), weak\_aug(X) } \\
        \vspace*{\baselineskip}
        \PyCode{$\mathtt {P_s}$ = $\mathtt {h_s}$($\mathtt {F_s}$($\mathtt {X_s}$))} \ \ \ \ \ \ \ \ \PyComment{NxC} \\
        \PyCode{$\mathtt {Z_t}$ = $\mathtt {F_{t}}$($\mathtt {X_t}$)}     \ \ \ \ \ \ \ \ \ \ \ \ \PyComment{NxC} \\

        \PyComment{topK nearest-neighbors lookup} \\
        \PyCode{$\mathtt {NN_t}$ = NN($\mathtt {Z_t}$, $\mathcal{S}$, topK)} \ \ \ \ \ \ \ \ \ \ \ \ \ \ \ \PyComment{(topKxN)xC}\\
        \PyCode{$\mathtt {Z_{topK}}$ = $\mathtt {Z_t}$.repeat(1, topK).reshape(-1, C)} \PyComment{(topKxN)xC} \\
        \colorbox[HTML]{ECF4FF}{\PyCode{$\lambda$ $\sim$  Uniform(0, 1)}} \\
        \colorbox[HTML]{ECF4FF}{\PyCode{$\mathtt {Z_{mix}}$ = $\lambda$*$\mathtt {NN_t}$ + (1-$\lambda$)*$\mathtt {Z_{topK}}$}} \\
        \vspace*{\baselineskip}
        \PyComment{$\mathtt {l_2-normalize}$} \\
        \PyCode{$\mathtt {P_{s\_norm}}$, $\mathtt {Z_{mix\_norm}}$, $\mathtt {Z_{t\_norm}}$ = normalize($\mathtt {P_s}$, $\mathtt {Z_{mix}}$, $\mathtt {Z_t}$, dim=1)} \\
        \PyComment{mm: matrix multiplication} \\
        \PyCode{$\mathtt {dist_{nn}}$, $\mathtt {mask}$ = 2-2*mm($\mathtt {P_{s\_norm}}$, $\mathtt {Z_{mix\_norm}}$), mask(N)} \PyComment{Nx(topKxN)}\\
        \PyCode{$\mathtt {dist}$ = 2-2*mm($\mathtt {P_{s\_norm}}$, $\mathtt {Z_{t\_norm}}$)} \ \ \ \ \ \ \ \ \ \ \ \ \ \ \ \ \ \ \PyComment{NxN}\\
        \colorbox[HTML]{ECF4FF}{\PyCode{$\mathtt {loss}$ = mm(dist, eye(N)).sum(dim=1) + mm($\mathtt {dist_{nn}}$, mask).sum(dim=1)/topK}}\\
        \vspace*{\baselineskip}
        \PyCode{loss.backward()} \\
        \PyComment{SGD update: student} \\
        \PyCode{update($\mathtt {F_s}$.params)} \\
        \PyComment{momentum update: $\mathtt {teacher}$} \\
        \PyCode{$\mathtt {F_{t}}$.params = $\mathtt {m}$*$\mathtt {F_{t}}$.params+(1-$\mathtt {m}$)*$\mathtt {F_s}$.params} \\
        \PyComment{update support set} \\
        \PyCode{enqueue($\mathcal{S}$, $\mathtt {Z_t}$)} \\       
        \PyCode{dequeue($\mathcal{S}$)} \\
        \vspace*{\baselineskip}
    \Indm 
    \PyCode{def mask(N):} \\
    \Indp   
        \PyCode{$\mathtt {mask}$ = eye(N).repeat(topK, 1).reshape(topK, N, -1))} \\
        \PyCode{return mask.permute(2, 1, 0).reshape(N, (topKxN))} \\
        \vspace*{\baselineskip}
    \Indm 
\caption{Pseudo-code of MNN (asymmetric) in a PyTorch-like style.}
\label{algo:algo-mnn}
\end{spacing}
\end{algorithm}
\end{minipage}
\end{center}

\section{Analysis of the mixture}\label{A2}
\begin{equation}
\begin{aligned}
Loss_x &= {\textstyle \sum_{i=0}^{K} w_i*||p^1 - \widetilde{z_i^2}||_2^2}\\
&=  \underset{(B.1-1)}{\underbrace{||p^1-z^2||_2^2}} + \frac{1}{K}{\textstyle \sum_{i=1}^{K} \underset{(B.1-2)}{\underbrace{||p^1 - \widetilde{z_i^2}||_2^2}}}\\
\end{aligned}
\label{eq:1}
\end{equation}

First, we rewrite Eq. (B.1-2) based on the $\widetilde{z_i^2} = \lambda *z_i^2 + (1-\lambda)*z^2$: 

\begin{equation}
\begin{aligned}
(B.1-2) &= ||p^1-\lambda * z_i^2 - (1-\lambda)*z^2||_2^2\\
 &= ||\lambda * (p^1 - z_i^2) + (1-\lambda)*(p^1 - z^2)||_2^2\\
 &= \lambda^2* \underset{(B.2-1)}{\underbrace{||p^1 - z_i^2||_2^2}} + (1-\lambda)^2* \underset{(B.2-2)}{\underbrace{||p^1 - z^2||_2^2}} + \underset{(B.2-3)}{\underbrace{2\lambda(1-\lambda) (p^1-z_i^2)^T(p^1-z^2)}}\\
 \end{aligned}
\label{eq:2}
\end{equation}

We ignore Eq. (B.2-3) to simplify Eq. (B.1-2):

\begin{equation}
\begin{aligned}
(B.1-2) &= ||p^1-\lambda * z_i^2 - (1-\lambda)*z^2||_2^2\\
 &\approx \lambda^2*||p^1 - z_i^2||_2^2 + (1-\lambda)^2*||p^1 - z^2||_2^2\\
 \end{aligned}
\label{eq:3}
\end{equation}

Finally, we can obtain a simplified version of $Loss_x$:

\begin{equation}
\begin{aligned}
Loss_x &= {\textstyle \sum_{i=0}^{K} w_i*||p^1 - \widetilde{z_i^2}||_2^2}\\
&= ||p^1-z^2||_2^2 + \frac{1}{K}{\textstyle \sum_{i=1}^{K}||p^1 - \widetilde{z_i^2}||_2^2}\\
&\approx ||p^1-z^2||_2^2 + \frac{1}{K}{\textstyle \sum_{i=1}^{K} \left \{ \lambda^2*||p^1 - z_i^2||_2^2 + (1-\lambda)^2*||p^1 - z^2||_2^2\right \}} \\
&= ||p^1-z^2||_2^2 + \frac{\lambda^2}{K}{\textstyle \sum_{i=1}^{K} ||p^1 - z_i^2||_2^2} +\frac{(1-\lambda)^2}{K}{\textstyle \sum_{i=1}^{K} ||p^1 - z^2||_2^2}\\
&=  \underset{(B.4-1)}{\underbrace{(1+(1-\lambda)^2)}} *||p^1 - z^2||_2^2 +  \underset{(B.4-2)}{\underbrace{\frac{\lambda^2}{K}}} *{\textstyle \sum_{i=1}^{K} ||p^1 - z_i^2||_2^2}\\
\end{aligned}
\label{eq:4}
\end{equation}

Since $\lambda \in [0, 1]$, we can observe $1 \le (B.4-1) \le 2$ as well as $0 \le (B.4-2) \le 1/K$. This implies that in MNN, the mixture operation further distinguishes the contribution of positive and neighbor samples to the model by the mixed coefficient $\lambda$. To delve into this effect, we fixed different values of $\lambda$ in MNN and conducted a detailed study of its impact on model performance. Table \ref{table: different_lamda} shows the linear classification results corresponding to different $\lambda$ values. Remarkably, we find that the model performs better when $\lambda$ takes a medium value.

\begin{table}[ht]
\centering
\setlength{\belowcaptionskip}{6 pt}
\caption{Linear evaluation results of MNN with varying $\lambda$ values.}
\begin{tabular}{ccccc}
\hline
\multicolumn{1}{c}{$\lambda$} & CIFAR-10       & CIFAR-100     & STL-10   & Tiny ImageNet  \\ \hline
$0.1$            & 90.46          & 67.25         & 91.38      & 49.15          \\
$0.3$            & 91.15          & 67.25         & 91.40      & 49.93          \\
$0.5$            & 91.27          & \textbf{67.56} & 91.21     & \textbf{50.31} \\
$0.7$            & \textbf{91.47} & 66.52          & \textbf{91.75}      & 49.26          \\
$0.9$            & 91.23          & 66.66          & 91.18      & 49.08          \\ \hline
\end{tabular}
\label{table: different_lamda}
\end{table}
\begin{figure}[!h]
	\centering
    \includegraphics[width=\linewidth]{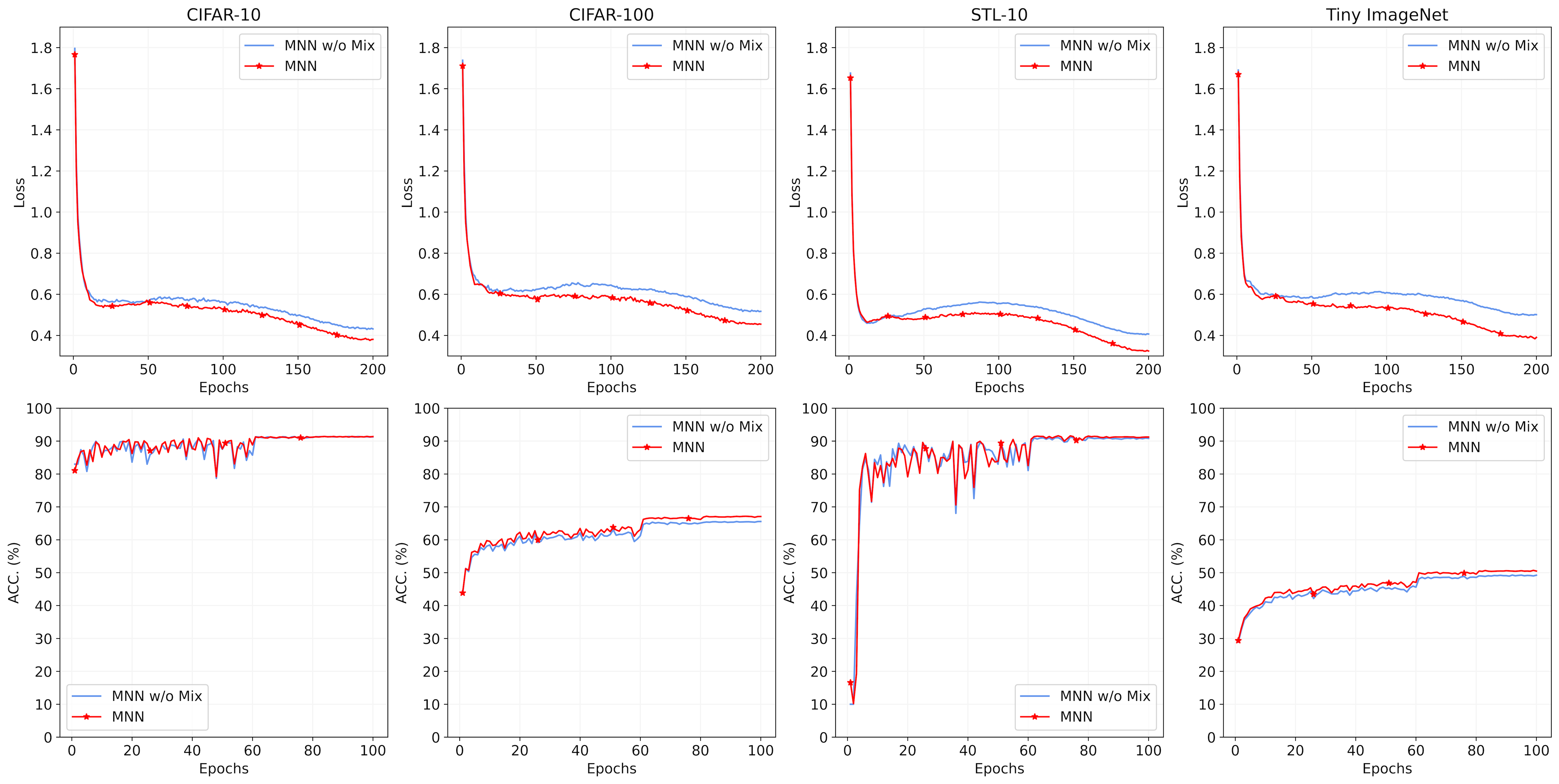}
	\centering
     \caption{The curves of the loss function and linear evaluation results on MNN and its variants without the mixture operation.}
     \label{figure: illustration_mnn_w_o_mix}
     \vspace{-1.0em}
\end{figure}
In addition, Fig.~\ref{figure: illustration_mnn_w_o_mix} illustrates the curves of the loss function values and linear evaluation results of the MNN with and without mixture. It can be observed from the figure that the mixture operation is effective in reducing the value of the loss function. This phenomenon may be due to the fact that the mixture operation helps to reduce the disturbance of the noise introduced by the false neighbors to the model.
\section{Inconsistency in the Cross-Attention Score}\label{A3}
\begin{figure}[!h]
	\centering
    \includegraphics[width=\textwidth]{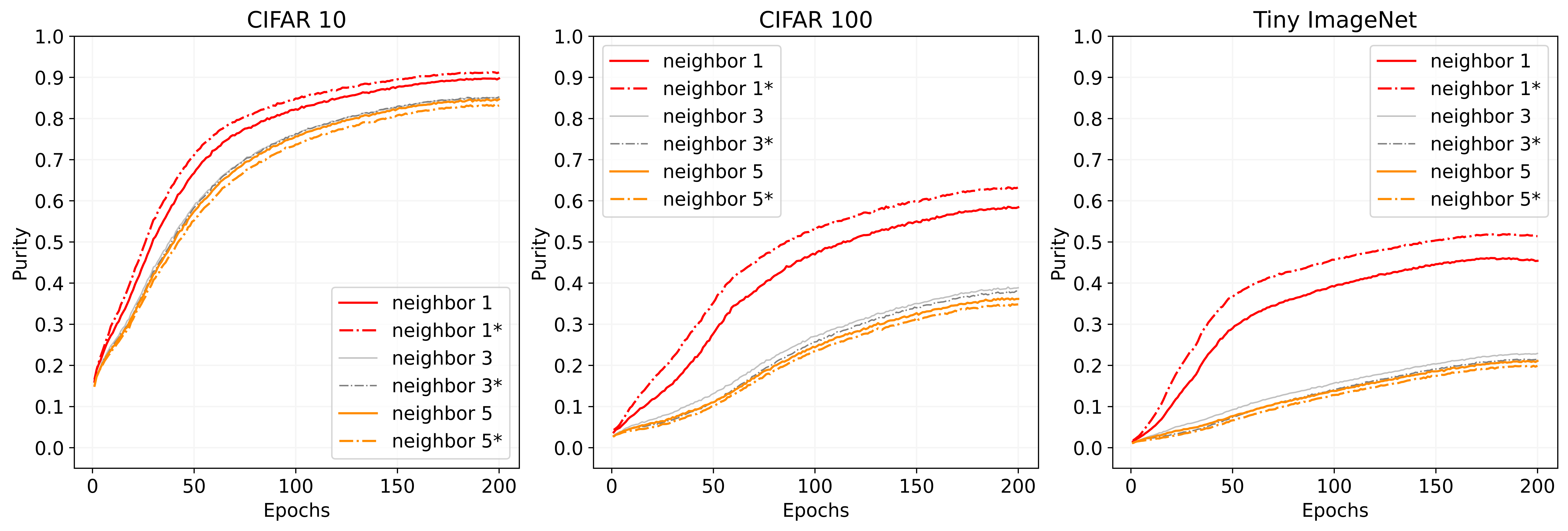}
	\centering
     \caption{Purity of different positions before and after reordering the set of neighbors using Cross Attention Scores. The $*$ symbol denotes the sorted purity. Notably, the sample in the 1st position at the front exhibits increased purity, while the sample in the 5th position at the back (in the default case with $K=5$) exhibits a corresponding decrease. This observation partially explains why the CAS improves the performance of the MSF. However, such modules may also assign higher weights to false neighbors due to the inconsistency in the CAS.}
     \label{figure: illustration_smsf_neighbor_purity}
     \vspace{-1.0em}
\end{figure}

In this section, we delve into the critical details of introducing the Cross-Attention Score to the MSF approach. When we rearrange the set of ordered neighbors ${\{z^2_i\}}_{i=1}^K$ based on the CAS, we obtain another ordered set ${\{z^{2'}_j\}}_{j=1}^K$. Fig.~\ref{figure: illustration_smsf_neighbor_purity} illustrates the purity at different neighbor positions, with * denoting the purity at positions 1, 3, and 5 in ${\{z^{2'}_j\}}_{j=1}^K$ (for visual clarity, we exclude the purity values at positions 2 and 4). Without *, it represents the purity at corresponding positions in the set ${\{z^2_i\}}_{i=1}^K$.

We observe that the CAS-induced inconsistency results in pushing samples with similar semantics to the positive embedding towards the front of the order, while false neighbors are placed towards the back. This phenomenon may contribute to the performance improvement seen when introducing the CAS to the MSF framework. However, the higher entropy of the weight distribution obtained by the CAS also suggests that it may assign larger weights to the false neighbors. This, in turn, can perplex the model during optimization since the weight assignments do not consistently reflect the actual similarity between samples.

\pagebreak
\noindent \textbf{Xianzhong Long} received his Ph.D. degree from the Department of Computer Science and Engineering at Shanghai Jiao Tong University in Shanghai, China. He is currently an associate professor at the School of Computer Science, Nanjing University of Posts and Telecommunications, Nanjing, China. His current research interests include self-supervised learning and adversarial machine learning.

\noindent \textbf{Chen Peng} received his B.S. degree in Computer Science and Technology from Nanjing University of Posts and Telecommunications. He is pursuing his M.S. degree in the School of Computer Science, at Nanjing University of Posts and Telecommunications, Nanjing, China. His current research interests include self-supervised learning, machine learning, and computer vision.

\noindent \textbf{Yun Li} received his Ph.D. degree in computer science from Chongqing University, Chongqing, China. He was a Post-Doctoral Fellow with the Department of Computer Science and Engineering, at Shanghai Jiao Tong University, Shanghai, China. He is a Professor at the School of Computer Science, Nanjing University of Posts and Telecommunications, Nanjing, China. He has published more than 60 refereed research papers. His current research interests include machine learning, data mining, and parallel computing.

\end{document}